\documentclass[10pt,twocolumn,letterpaper]{article}

\usepackage{iccv}
\usepackage{times}
\usepackage{epsfig}
\usepackage{graphicx}
\usepackage{amsmath}
\usepackage{amssymb}

\usepackage[textsize=tiny]{todonotes}

\usepackage{wrapfig,lipsum,booktabs}

\usepackage{bbding}
\usepackage{array}
\usepackage{arydshln}

\usepackage{subfigure}
\usepackage{adjustbox}
\usepackage{xcolor}
\usepackage{threeparttable}
\usepackage{makecell}
\usepackage{listings}
\usepackage{pifont}%

\usepackage[noblocks]{authblk}

\newcommand{\aka}{\emph{a.k.a}}

\newcommand{\red}[1]{\textcolor{black}{#1}}

\usepackage{algpseudocode}
\usepackage[linesnumbered,ruled]{algorithm2e} 
\usepackage{booktabs} 

\usepackage{hyperref}
\hypersetup{pagebackref=true,letterpaper=true,breaklinks=true,bookmarks=false, colorlinks, linkcolor=blue,  anchorcolor=blue,citecolor=blue,anchorcolor=blue,urlcolor=black }

\iccvfinalcopy 

\begin{document}

\title{GlueGen: Plug and Play Multi-modal Encoders for X-to-image Generation}

\author{Can Qin$^\star$\thanks{This work was done when Can Qin interned at Salesforce AI Research. Primary contact: qin.ca@northeastern.edu}, 
Ning Yu$^\dagger$,
Chen Xing$^\dagger$,
Shu Zhang$^\dagger$,
Zeyuan Chen$^\dagger$, \linebreak
Stefano Ermon$^\ddagger$,
Yun Fu$^\star$, 
Caiming Xiong$^\dagger$,
Ran Xu$^\dagger$ \linebreak
$^\star$Northeastern University, $^\dagger$Salesforce AI Research, $^\ddagger$Stanford University \linebreak
\small{\texttt{qin.ca@northeastern.edu,  ermon@cs.stanford.edu, yunfu@ece.neu.edu, \linebreak \{ning.yu, cxing, shu.zhang, zeyuan.chen, cxiong, ran.xu\}@salesforce.com }}
}

\maketitle
\pagestyle{empty}
\ificcvfinal\thispagestyle{empty}\fi

\begin{abstract}
Text-to-image (T2I) models based on diffusion processes have achieved remarkable success in controllable image generation using user-provided captions. However, the tight coupling between the current text encoder and image decoder in T2I models makes it challenging to replace or upgrade. Such changes often require massive fine-tuning or even training from scratch with the prohibitive expense.
To address this problem, we propose GlueGen, which applies a newly proposed GlueNet model to align features from single-modal or multi-modal encoders with the latent space of an existing T2I model. The approach introduces a new training objective that leverages parallel corpora to align the representation spaces of different encoders. Empirical results show that GlueNet can be trained efficiently and enables various capabilities beyond previous state-of-the-art models: 1) multilingual language models such as XLM-Roberta can be aligned with existing T2I models, allowing for the generation of high-quality images from captions beyond English; 2) GlueNet can align multi-modal encoders such as AudioCLIP with the Stable Diffusion model, enabling sound-to-image generation; 3) it can also upgrade the current text encoder of the latent diffusion model for challenging case generation. By the alignment of various feature representations, the GlueNet allows for flexible and efficient integration of new functionality into existing T2I models and sheds light on X-to-image (X2I) generation.\footnote{Code will be available at: \href{https://github.com/salesforce/GlueGen}{https://github.com/salesforce/GlueGen}}

\end{abstract}

\section{Introduction}

Text-to-image (T2I) generative models have made great progress in the last few years thanks to algorithmic advances and the availability of large-scale paired training datasets
~\cite{ramesh2022hierarchical,yu2022scaling,rombach2022high,schuhmann2021laion,schuhmann2022laion}. Diffusion-based T2I generative models in particular 
have achieved remarkable results in terms of image quality
~\cite{ho2022classifier,nichol2021glide,balaji2022ediffi,saharia2022photorealistic,mou2023t2i,zhang2023adding}. 
Despite these strong results, controllable generation for these methods is still challenging: generated images are often not faithful to the captions, compositional capabilities are lacking, and prompt engineering is often required to achieve the desired results~\cite{dall-e-prompt-book}. Moreover, most large-scale models have only been trained on English text captions, greatly limiting their use across the world.

\begin{figure}[t]
\vspace{-2mm}
\includegraphics[width=1\linewidth]{   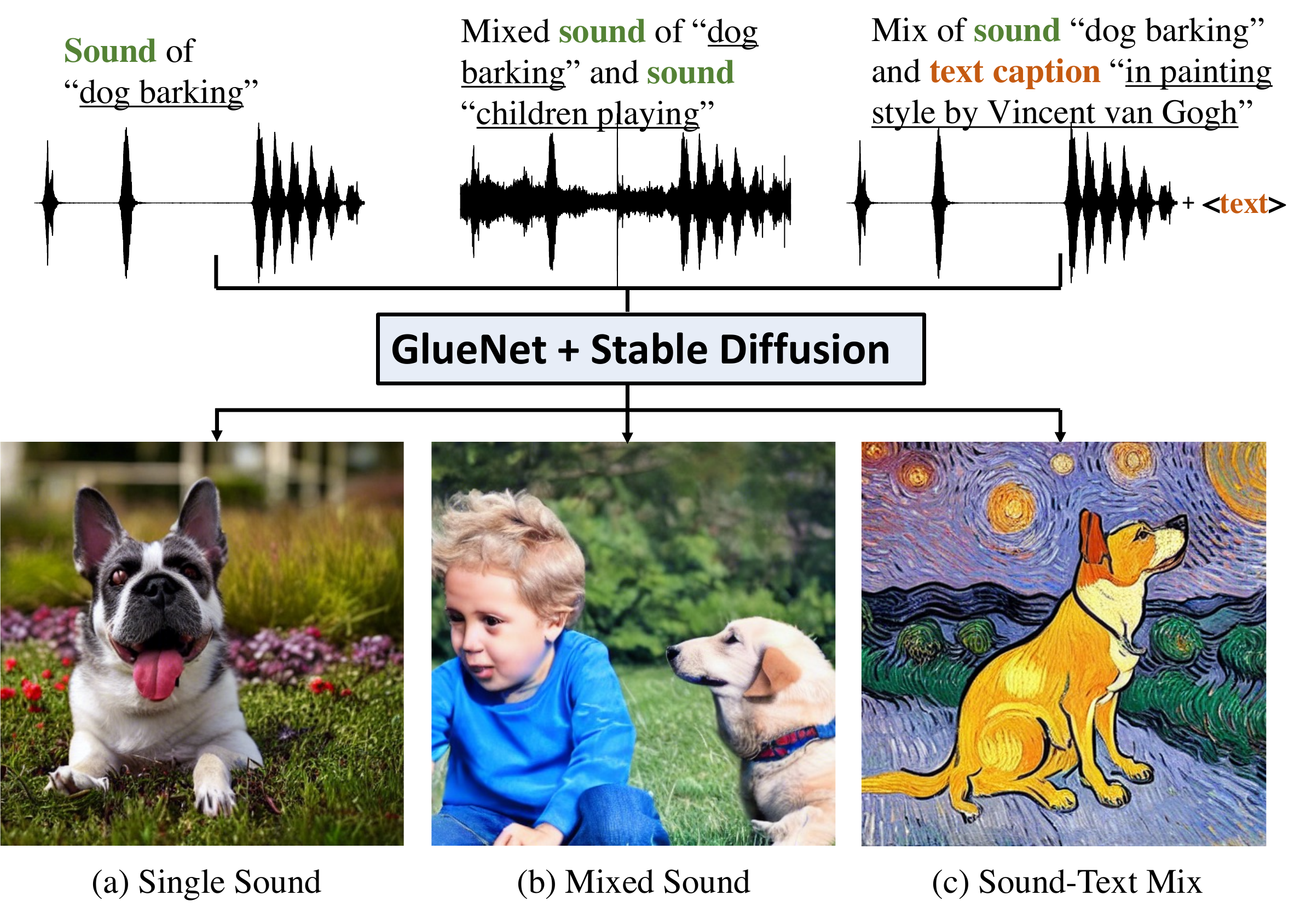}
\caption{Setting of GlueGen. GlueNet is trying to provide an adaptable portal for the Stable Diffusion model to input multi-modal data, such as text, audio, \ie, (a) and (b), or text-audio hybrid signals, \ie (c), for X-to-image generation.}\label{fig:demo-resuls}
\end{figure}

Recent research has emphasized the crucial role of text encoders in improving Text-to-Image (T2I) models' performance, and their ability to comprehend and represent text is considered a bottleneck for image generation~\cite{saharia2022photorealistic, croitoru2022diffusion}. However, the current T2I models' text encoders are often trained on short image captions, which limits their performance on complex prompts and challenges their quality of feature extraction~\cite{rombach2022high}. Furthermore, T2I models' capacities are limited to generating images from text, and they cannot incorporate multimodal conditions such as sound and audio easily. Nevertheless, replacing the text encoder in existing T2I models is challenging since the text encoder and image generator's representation spaces are tightly coupled~\cite{rombach2022high, ramesh2022hierarchical}. This severe domain gap between the new conditions and the existing model impedes the image generation's final performance, and training the entire T2I model from scratch, with higher quality image-caption pairs, would be prohibitively expensive~\cite{edwards_2022}.\footnote{The cost of training a Stable Diffusion model is around 600K USD.}

As seen in Fig.~\ref{fig:demo-resuls}, we propose GlueNet to address the challenge of efficiently replacing or upgrading the text encoder in existing diffusion-based T2I models. With GlueNet, off-the-shelf pre-trained language models and multimodal encoders can be easily aligned with image encoders of T2I models, greatly enlarging their functionalities at a low cost. Importantly, this can be achieved without requiring retraining from scratch or even finetuning, maintaining the representation alignment between the text and image encoders. The proposed method follows an encoder-decoder structure. The encoder of GlueNet first aligns the representation space of the new condition encoder with that of the T2I model's image generator, minimizing both element-wise and global-wise discrepancy. Then, the decoder of GlueNet maps the aligned condition representations back to the original representation space of the new condition encoder by minimizing the reconstruction loss, preserving rich semantics captured by the pre-trained model during alignment training. Align existing models would inevitably decrease feature discriminability~\cite{pmlr-v97-chen19i,cui2020towards}, which makes the feature decoder necessary.
The entire training of GlueNet requires only a parallel corpus with the same content but different modalities or languages. At inference time, only the encoder of GlueNet is applied on top of the new condition encoder for representation alignment.

To verify the effectiveness of the proposed framework, we conducted three major experiments ranging from single- and multi-modal encoders. Firstly, we upgraded the existing text encoders of the Latent Diffusion Model~\cite{rombach2022high} using a stronger language model, T5-3B~\cite{raffel2020exploring}. Our model showed competitive improvements in FID score and user study ranking compared to the baselines but it still required finetuning for the overall performance boost. Secondly, we aligned a multilingual language model, XLM-Roberta-L~\cite{conneau2019unsupervised}, using our approach, enabling multilingual text-to-image generation. It achieved competitive results of translation-based models under a significantly lower training cost. Finally, we demonstrated GlueNet's capability to bring new functionalities beyond text signals into existing T2I models. 
The alignment of the AudioClip~\cite{guzhov2022audioclip} encoder enables sound-to-image generation without requiring any parameter finetuning of the image generator. This new capability allows the existing Stable Diffusion model to generate high-quality images that correspond to sound signals such as dogs barking and street music. This new capability goes beyond the traditional T2I generation and opens up new possibilities for creating multimedia content towards X-to-Image (X2I) generation.

\begin{figure*}[t]
\begin{center}
\includegraphics[height=0.3\linewidth,width=\linewidth]{   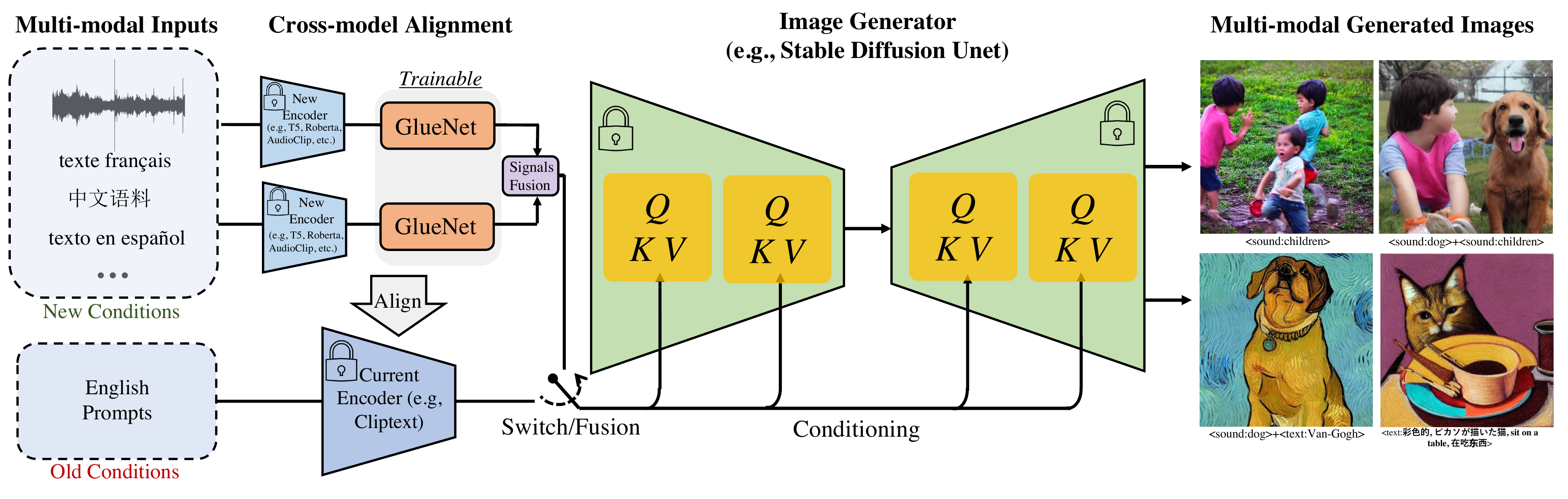}
\end{center}
\caption{
Illustration of our desired GlueGen framework. With the proposed GlueNet model of the GlueGen framework, the pre-trained image generator (\ie UNet) can be bridged to off-the-shelf single- or multi-modal encoders to expand their functionalities, \ie, multilingual/sound-to-image generation, within a limited budget. GlueNet is trained offline and does not require back-propagation of UNet and image-text pairs for training. Therefore, GlueGen is flexible and efficient to achieve.
}
\label{fig:setting}
\end{figure*}

Our contributions can be summarized as follows:

\begin{itemize}
\item To the best of our knowledge, this is the first work to consider the problem of efficiently aligning a pre-trained audio model with a pre-trained T2I diffusion model for sound-to-image generation.

\item  Extensive experiments on text-to-image generation benchmarks demonstrate the superiority of our model over the baseline LDM method on both image quality and language controllability.

\item Our framework also enables text-to-image generation beyond English prompts without the need of multilingual image-text pairs for retraining.

\end{itemize}

\section{Related Works}

\subsection{Text-to-Image Models}

Generative Adversarial Nets (GANs) \cite{Goodfellow2014GAN} is one of the major frameworks in text-to-image generation. The method in \cite{reed2016generative} was an early stage approach to use a GAN to train a text-to-image generation network. Since then, other GAN based methods, \eg,   Stack-GAN \cite{han2017stackgan}, Attn-GAN \cite{xu2018Attngan} and SD-GAN \cite{Yin2019SDGAN}, have obtained promising results. In addition, recent works showed more improvements on the generation quality. DM-GAN \cite{zhu2019dmgan} improved text-to-image performance by introducing a dynamic memory component. DF-GAN \cite{tao2022dfgan} designed a fusion module to fuse text and image features. LAFITE \cite{zhou2021lafite} took advantage of CLIP's model to construct pseudo image-text pairs, and proposed a GAN model to learn text-image pairs.

Auto-regressive transformer is another major framework in text-to-image generation. DALL-E \cite{ramesh2021zero} and CogView \cite{ding2021cogview} adopted an autoregressive transformer \cite{Vaswani2017Transformer} to train the correspondences between visual tokens and text tokens. Parti \cite{yu2022scaling} used a powerful visual encoder, ViT-VQGAN \cite{Yu2022VitVQGAN}, to improve results upon DALL-E. The method in \cite{gafni2022make} is similar to CogView and DALL-E, while introducing additional controlling elements to improve the tokenization process. N\"{U}WA \cite{wu2022nuwa} and NUWA-infinity \cite{wu2022nuwainfinity} trained autoregressive visual synthesis model to support both text-to-image and text-to-video generation tasks.

Concurrently, diffusion models \cite{sohl2015deep,song2019generative,song2020score} became a main research focus. In such methods, noise is added to an image, and a score network is trained to denoise and recover the input image. GLIDE \cite{nichol2021glide} performed guided inference with and without a classifier network to generate high-fidelity images. DALL-E 2 \cite{ramesh2022hierarchical} and Imagen \cite{saharia2022photorealistic} set new state-of-the-art results in the text-to-image generation. In DALL-E 2, a prior to produce CLIP image embeddings conditioned on text was learned, and a diffusion model was used to decode the image embeddings to an image. In Imagen, a large-scale frozen text encoder T5-XXL \cite{raffel2020exploring} was adopted to generate embeddings, and a cascade of conditional diffusion models was used to map these embeddings to images of increasing resolutions. Latent diffusion model and stable diffusion model \cite{rombach2022high} are state-of-the-art methods that apply the diffusion model on the latent embedding space as in \cite{sinha2021d2c,vahdat2021score}. These methods are computationally friendly while achieving impressive results. To incorporate more conditions for controllable content generation, \cite{zhang2023adding,mou2023t2i,brooks2022instructpix2pix} introduced additional parameters with new data to inject external knowledge within the current pre-trained model. \cite{chen2022altclip} tried to alter the language model for more capacities. However, all these models are designed to adapt text and image (or similar 2d grid) conditions without the attempt for sequential signals such as audio.

\subsection{Sound-to-Image Generation}

Sound and image are two distinct modalities. \cite{lee2022sound} attempts to address sound-guided semantic image manipulation by aligning sound features with a GAN model. CoDI~\cite{chang2023design} take a two-stage strategy which pre-trains individual diffusion model with aligned prompt encoder in the beginning. Then it aligns latent to project multimodal features into a shared space. AAI~\cite{yang2023align} introduces sound-guided image generation, editing and stylization into one model.

\section{Preliminary}
\subsection{Latent Diffusion Model}
The Latent Diffusion Model (LDM)~\cite{rombach2022high} and its extension, \ie, Stable Diffusion\footnote{https://github.com/CompVis/stable-diffusion}, which are variants of Denoising Diffusion Probabilistic Model (DDPM)~\cite{ho2020denoising} family, are selected as our baseline models due to their excellent balance in efficiency and visual quality. In general, LDM is composed of two stages. Firstly, there is an auto-encoder (AE) pre-trained by enormous images with the regularization in either KL-divergence or vector quantization~\cite{van2017neural,agustsson2017soft}. An encoder network, \ie, $E$, is applied for latent feature extraction as $z=E(x)$, which can be mapped back to image space by a decoder network. The input images $x$ and reconstructed images $\hat{x}$ are almost identical $x \approx \hat{x}$. 

Secondly, LDM trains a diffusion model in the latent space~\cite{sinha2021d2c,vahdat2021score}. It follows the standard DDPM~\cite{ho2020denoising} with denoising loss and uses U-net~\cite{ronneberger2015u} as the image decoder as in ~\cite{song2019generative}. To enable generative controllability, LDM has applied multiple conditioning signals ($y$) such as text, mask, or layout, encoded aside and injected into the U-net, with the help of cross-attention layers. This can be formulated as:
\begin{equation}
    \mathcal{L}_{LDM} := \mathbb{E}_{z\sim E(x), \varepsilon \sim N(0,1), t, y} \left [ \left \| \varepsilon - \varepsilon_{\theta}(z_t,t,c_{\phi}(y)) \right \| \right ],
\end{equation}
where $t$ represents the time step, $z_t$ is the noise corrupted latent tensor at time step $t$, and $z_0 = E(x)$. $\varepsilon$ is the unscaled Gaussian noise, $c_{\phi}$ is the conditioning network parameterized by $\phi$ and $\varepsilon_{\theta}$ is the Unet-like denoising network (\aka, image decoder). The parameters of both conditioning and denoising networks, \ie, ${\theta, \phi}$, are trained by the LDM loss. During inference,  clean images can be generated via  classifier-free guidance~\cite{ho2022classifier} as:
\begin{equation}
\hat\varepsilon_{\theta}(z_t|y) = \varepsilon_{\theta}(z_t) + s \cdot (\varepsilon_{\theta}(z_t, c_{\phi}(y)) - \varepsilon_{\theta}(z_t)),
\label{eq:sampling}
\end{equation}
where $s$ is the guidance weight to balance text controllability and image fidelity. 

The basic T2I LDM model is trained on Laion-400M~\cite{schuhmann2021laion} dataset. The Stable Diffusion Model (SDM) was trained with more epochs and training data, \ie, Laion2B-en and Laion-high-resolution~\cite{schuhmann2022laion}. 

\begin{figure*}[t]
\begin{center}
\includegraphics[height=0.28\linewidth,width=\linewidth]{   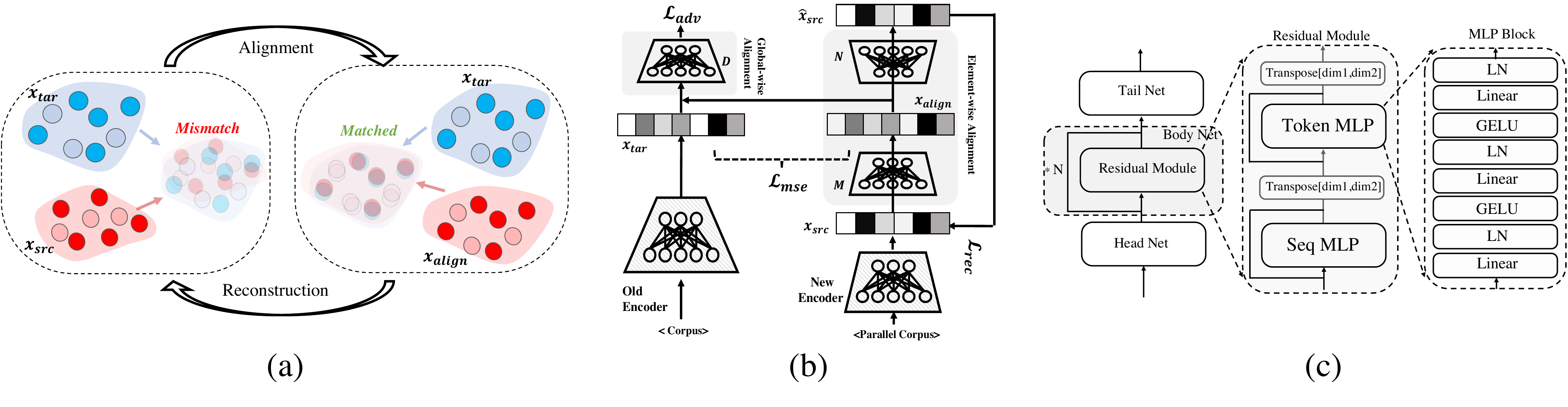}
\end{center}
\vspace{-6mm}
\caption{(a) Illustration of features transformation throughout the model translation/alignment. (b) The general pipeline and learning objectives of our proposed GlueNet. (c) Detailed architecture of GlueNet Encoder/Decoder.
}
\vspace{-4mm}
\label{fig:method}
\end{figure*}

\subsection{Condition Encoder}
LDM uses  Bert-like~\cite{devlin2018bert} encoder as the text encoder jointly trained with the image decoder using the DDPM loss. SDM uses a pre-trained CLIP~\cite{radford2021learning} text encoder frozen during training. It has been found that increasing the size of the text encoder is very helpful for performance~\cite{saharia2022photorealistic}. In order to upgrade existing models, our goal is to be able to efficiently plug in more advanced language condition encoders such as T5-3B~\cite{raffel2020exploring}, AudioClip~\cite{guzhov2022audioclip}, or XLM-Roberta~\cite{devlin2018bert}.

\section{GlueNet}
\label{sec:method}

The text encoder is one of the key components of T2I models, as generation requires precise and fine-grained text embedding for guidance. The overall performance can be greatly boosted by increasing the size of the text encoder~\cite{saharia2022photorealistic}, however, upgrading an existing text model to a more powerful one is challenging. 
Because existing T2I models are not modular, 
directly replacing (or augmenting with) another condition encoder does not work. The key technical challenge is the mis-alignment between the new condition encoder and the old image decoder. Moreover, joint finetuning also falls short due to catastrophic forgetting occuring when updating well trained parameters. 

Considering these two different condition encoders as source and target domains, we apply a neural network to learn to align. As shown in Fig.~\ref{fig:setting}, this paper has presented a general framework called GlueGen.  Within the framework, GlueNet works as an additional module to bridge the new language model and the old image decoder.

\subsection{Objectives}

\red{Given sets of parallel corpus $\mathcal{X}_{src} = \{x_{ij}^s|i=1,...,L^s, j=1,...,N^s\}$  and $\mathcal{X}_{tar} = \{x_{ij}^t|i=1,...,L^t, j=1,...,N^t\}$ with $L$ as length of tokens and $N$ as total quantity,  the source model ${F}^{s}$ and a target encoder ${F}^{t}$ have mapped the raw data into embeddings denoted as $\mathcal{S} = \{s_{ij}|i=1,...,L^s, j=1,...,N\}$ where $s_{ij} = F^{s}(x_{ij}^s) \in \mathbb{R}^{d_{s}}$ and  $\mathcal{T} = \{t_{ij}|i=1,..,L^t, j=1,...,N\}$ where $t_{ij} = F^{t}(x_{ij}^t) \in \mathbb{R}^{d_{t}}$. Due to the different models, there is severe distribution mismatch between the source and target features $p(s) \neq p(t)$.}
The  GlueNet is an autoencoder whose encoder $M(\cdot, \Theta_{M})$ learns to map the new features from a source language model, \red{\ie, $s \in \mathcal{S}$,} to align with the current target encoder, \red{\ie, $t \in \mathcal{T}$, where $p(M(s, \Theta_{M})) \approx p(t)$}. \red{Therefore,} this translation (\ie, ${F}^{s} \overset{\tiny{\Theta_{M}} }{\longrightarrow}
 {F}^{t}$ ) enables the image generator to understand the new  features coming from the new condition model without finetuning. To achieve this aim, we consider it as a domain adaptation~\cite{ganin2016domain,qin2019pointdan,long2018conditional} problem and apply both element-wise and distribution-wise alignment losses, which includes the minimization of the mean square error (MSE) loss, \ie, $\mathcal{L}_{mse}$, and the adversarial loss, \ie, $\mathcal{L}_{adv}$, measured over a discriminator network, \ie, $D(\cdot, \Theta_D)$:
\begin{align}
\mathcal{L}_{mse}( \Theta_M) & := \mathbb{E}_{x_t \sim X_t, x_s  \sim X_s}[||{F}^{t}(x_t)- \notag \\  & M(F^{s}(x_s), \Theta_M)||_{2}^{2}], \label{eq:loss-mse} \\
\mathcal{L}_{adv}( \Theta_D, \Theta_M) & :=  \mathbb{E}_{x_t  \sim X_t}[\log{D({F}^{t}(x_t), \Theta_D)}] \notag \\ + & \mathbb{E}_{x_s  \sim X_s}[\log{1-D(M({F}^{s}(x_s), \Theta_M), \Theta_D)}], \label{eq:loss-adv}
\end{align}
where $X_s$ and $X_t$ denote the source and target inputs respectively and they can be the same prompts in English (\eg, LDM $\rightarrow$ T5-3B~\cite{raffel2020exploring}) or in bilingual but parallel content for multilingual T2I or audio-label pairs.

Some insightful investigations in transfer learning reveal that the cross-domain alignment will inevitably decrease feature discrimination~\cite{pmlr-v97-chen19i,cui2020towards}. To project the rich semantics necessary for model upgrading, we further apply a decoder network, \ie,  $N(\cdot, \Theta_N)$, for feature reconstruction:
\begin{align}
    \mathcal{L}_{rec}&(\Theta_M, \Theta_N)  :=  \notag \\ &\mathbb{E}_{x_s  \sim X_s} [||x_s- N(M(F^{s}(x_s), \Theta_M), \Theta_N)||_{2}^{2} ]. \label{eq:loss-rec}
\end{align}

 GlueNet is trained in an end-to-end manner and the details of which  can be referred to in Appendix and Sec.~\ref{sec:experiments}. \cite{ma2022principles} introduced that a universal learning engine should seek a compressive closed-loop transcription with good property in structure-preserving. Our cross-model mapping also requires a similar functionality.  GlueNet desires stability after a loop whose input and output should keep almost equivalent, \ie, $x \approx \hat{x} \approx \hat{\hat{x}} \approx ... $, due to the constraint of reconstruction loss. Therefore, inputting the reconstructed data $\hat{x}$ to  GlueNet is expected to be consistent with the previous loop. This is helpful for accelerating  GlueNet training for T5-to-LDM adaptation.

\subsection{Model Architecture}
\label{sec:arch}
The input data to alignment can be represented as the sequence of tokens: $\mathrm{X}_{i,*} = \{x_{i,1}, x_{i,2}, ...,x_{i,L}\}$ where $\mathrm{X}_{i,*} \in \mathcal{X}$ and ${ x_{i,j} \in \mathbb{R}^{C}}$, with sequence length $L$ and token dim $C$. Inspired by the methods of image super-resolution~\cite{lim2017enhanced,ahn2018fast} to densely regress target data under different input-output dimensions, the translator network stacks three sub-nets in sequence, as seen in Fig.~\ref{fig:method} (c). The head net is only used for simple feature transformation. The body net has many residual modules for fine-grained representation learning. The tail network is appended and is mainly employed for dimension conversion to match the target tensors. Every residual module is implemented based on the MLP-based token mixer~\cite{tolstikhin2021mlp,ma2021rethinking} with high efficiency and strong ability in representation learning. Within the MLP-mixer block, it consists of a Token Mixer and a Sequence Mixer for orthometric purposes. The Token Mixer learns the representation of each token with shared MLP, and the Sequence Mixer is designed for learning the same channel in sequential tokens:
\begin{align}
 \mathrm{U}_{\cdot,i} = & \mathrm{X}_{\cdot,i} + \mathrm{W}_{2}\sigma(\mathrm{W}_{1} * \mathrm{LN}_{\gamma_1, \beta_1}(\mathrm{X})_{\cdot,i}), i=1,...,C \\
  \mathrm{X}_{j,\cdot} = & \mathrm{X}_{j,\cdot} + \mathrm{W}_{4}\sigma(\mathrm{W}_{3} * \mathrm{LN}_{\gamma_2, \beta_2}(\mathrm{U})_{j,\cdot}),  j=1,...,L
\end{align}
where $\mathrm{X}_{\cdot,\cdot}$ is the feature, $\mathrm{LN}_{\gamma, \beta}$ denotes layer normalization parameterized by $\gamma$ and $\beta$, and $\sigma$ indicates the non-linear operator such as GELU~\cite{hendrycks2016gaussian}. The $\mathrm{U}_{\cdot,i}$ is the outputs of Seq MLP and $\mathrm{W}$ represents the weight matrix within the MLP mixer.

\subsection{Cross-modality Alignment}
\label{sec:cross-modal}
\begin{figure}[t]
\vspace{-2mm}
\includegraphics[width=1\linewidth]{     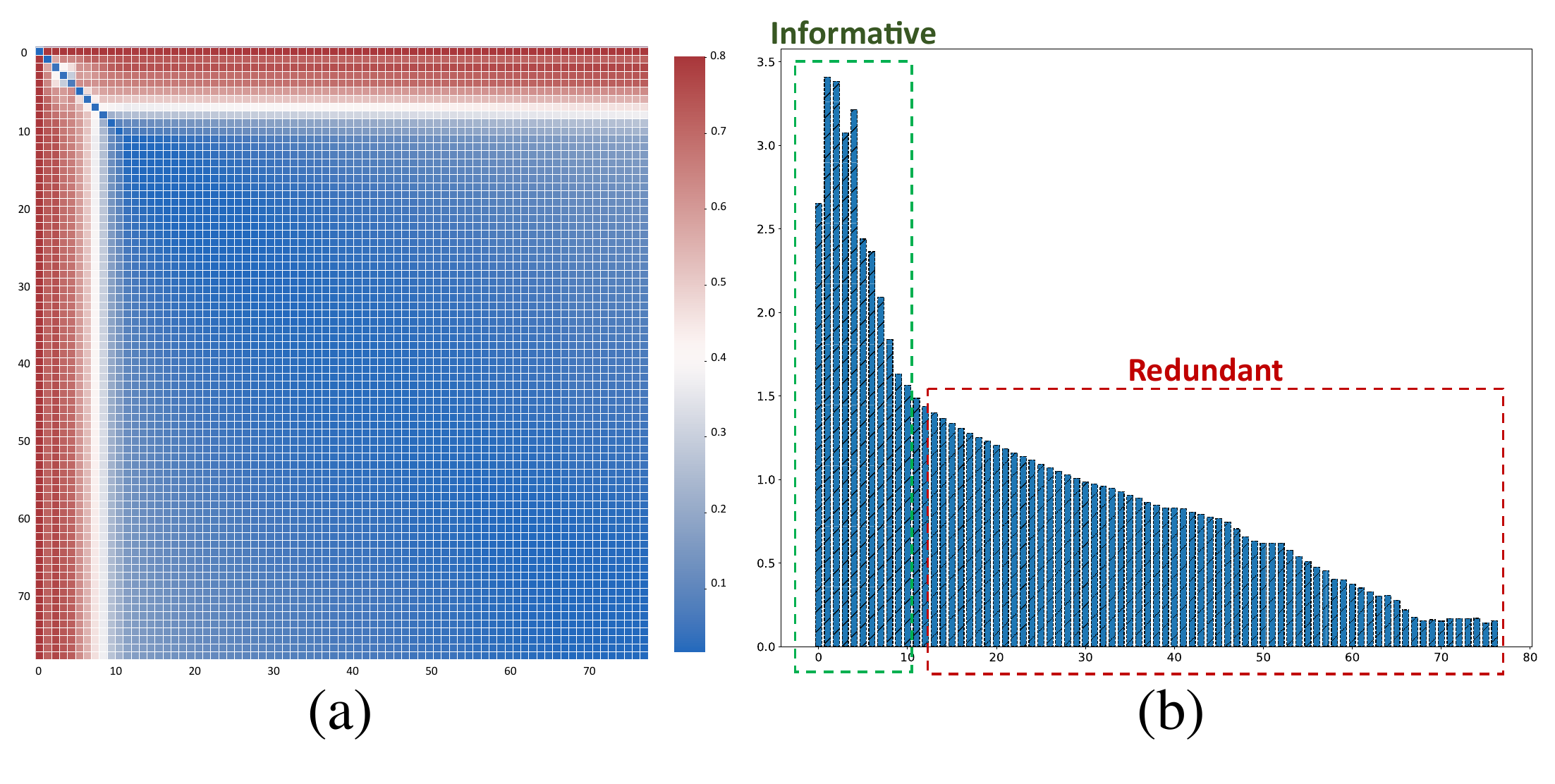}
\vspace{-7mm}
\caption{Signal analysis of CLIPText encoder on 77 tokens (used for Stable Diffusion) averaging over 10,000 randomly sampled prompts. (a) Average dis-similarity map (1-cosine) of CLIPText tokens. (b) The distance between each token and the last token, \ie, $dist(s_{\cdot,j}, s_{\cdot,L})$ with $L$ tokens. We can conclude that only the top K tokens are informative.  }\label{fig:signal-clip}
\vspace{-4mm}
\end{figure}
GlueNet provides a way to align the cross-modal representations between the text encoder, CLIPText~\cite{radford2021learning}, with the AudioClip~\cite{guzhov2022audioclip} model that maps sound signals into embedding spaces. This alignment is useful for extending the functionality of diffusion models like Stable Diffusion, which only takes text signals as input. However, achieving this alignment is challenging because the length of the tokens in each model is different. CLIPText's embeddings require 77 tokens in a sequence, while AudioClip maps sound signals to a single token with a sequence length of 1. Using the standard GlueNet directly fails in this case.

During our exploration, we found that condition information is not uniformly distributed across all 77 tokens of CLIPText outputs. The top K tokens contain the majority of the information. We computed an average dissimilarity map over the CLIPText tokens, visualized in Fig.~\ref{fig:signal-clip} (a), and observed a prominent gap around the eighth or ninth token, with the later ones being highly similar. We also found that corrupting the last token embedding with some random noise would not degrade the generative results. Therefore, it can be concluded that the contribution of different tokens decreases progressively and could be estimated by the feature distance with the last token, $w_{\cdot,j} = dist(s_{\cdot,j}, s_{\cdot,M})$, as shown in Fig.~\ref{fig:signal-clip} (b). We take the l2 distance as $dist(\cdot, \cdot)$ and use this value to reweight the GlueNet objective over different tokens for the sound-to-image generation.

Cross-modal signals fusion is another contribution. To do this, we select the top K tokens of each modality output from GlueNet. Then, the average of the remaining later tokens (excluding last K tokens) of two modality features is concatenated with the top ones. This is a non-parametric operation without any training.  K is empirically chosen from 4 to 8 (or higher). For more details, please refer to Appendix. 


\begin{figure*}[t]
\centering
\includegraphics[height=0.25\linewidth,width=\linewidth]{   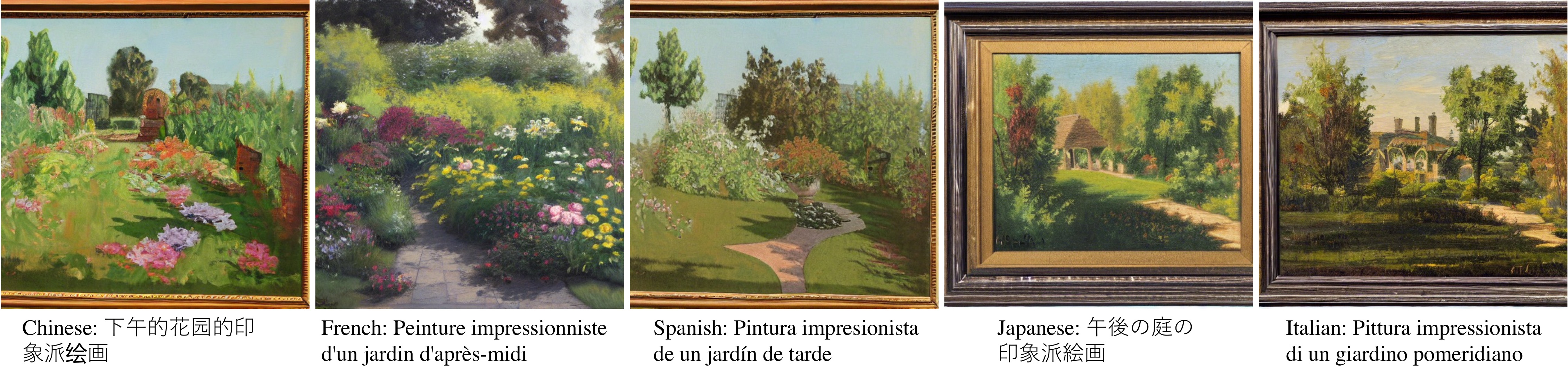}
\vspace{-7mm}
\caption{Multilingual generation results in resolution 512 $\times$ 512  of XLM-Roberta + Glue-Net + SDM decoder (sd-v1-4) with the same caption,
\ie, ``afternoon garden oil painting painted by impressionists". With the help of different Glue-Nets and multilingual text encoder, the SDM decoder can support different languages including Japanese, Italian, Chinese, French and Spanish. The guidance weight is assigned as 7.5 and PLMS~\cite{liu2022pseudo} sampling steps are 50. }\label{fig:multi}
\vspace{-3mm}
\end{figure*}

\section{Experiments}
\label{sec:experiments}
\subsection{Experiments Setup}
\label{sec:experiments-setup}

Upon the shoulders of giants, this paper applies the Latent-Diffusion Model (LDM) and Stable Diffusion Model (SDM)~\cite{rombach2022high} as the baselines with a similar model but different text encoders and scalabilities. LDM uses a Bert-like~\cite{devlin2018bert} model as the text encoder that is jointly updated with the image decoder based on DDPM loss, while SDM uses a frozen pre-trained CLIP text encoder  during training. The LDM is designed for the images in 256 $\times$ 256 and SDM targets 512 $\times$ 512 images. We implement the multilingual T2I and sound-to-image generation upon the SDM. 

\begin{wrapfigure}{r}{0.6\linewidth}
\includegraphics[width=\linewidth]{   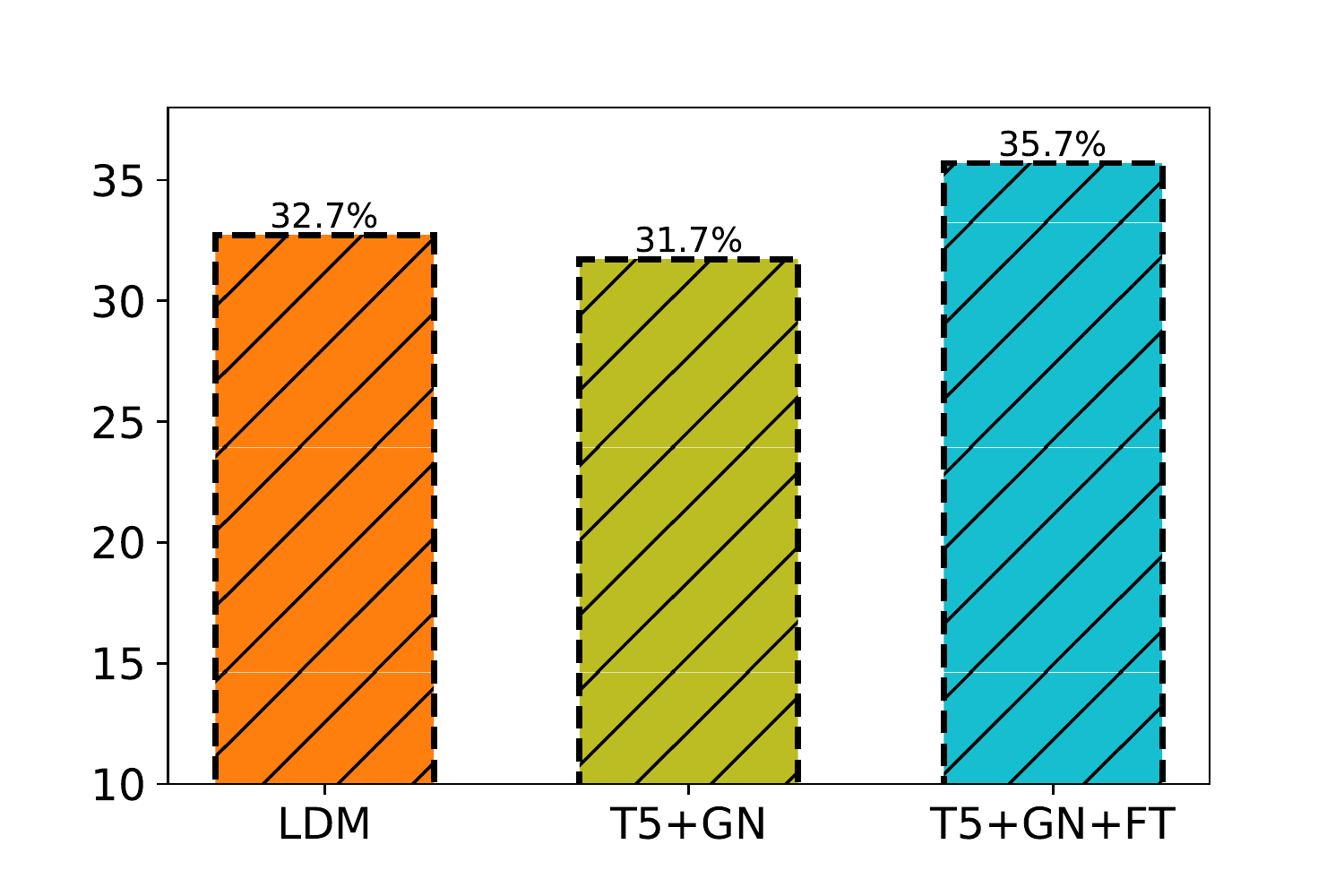}
\vspace{-2mm}
\caption{Percentage of user votes for text controllability with the prompts from Drawbench~\cite{saharia2022photorealistic} and Winoground~\cite{thrush2022winoground} for T5-to-LDM.}\label{fig:user-study}
\vspace{-2mm}
\end{wrapfigure}

\textbf{Implementation.} The GlueNet is implemented as the Fig.~\ref{fig:method} with an encoder $M$ and a decoder $N$ in symmetrical architecture consisting $\sim$51M parameters. For simplicity in denotation, we refer GlueNet encoder as GlueNet in the following experiments, whose body net consists of 5 residual modules as default. We take the AdamW~\cite{loshchilov2017decoupled} as the optimizer based on PyTorch~\cite{NEURIPS2019_9015}. The learning rate is assigned as 1$\times 10^{-4}$ for GlueNet training and 1$\times 10^{-6}$ for LDM-T5-GlueNet finetuning. The GlueNet is trained on Nvidia-A100, requiring 0.5 to 5 GPU days for different uses.

\textbf{Condition Encoders.} We apply the T5-3B~\cite{raffel2020exploring} to upgrade the text encoder of LDM Bert. T5-3B has 6x more parameters than LDM Bert and is trained by an enormous text corpus beyond the image captions. Furthermore, to enable multilingual T2I generation, we choose the XLM-Roberta-L~\cite{devlin2018bert} as the new condition encoder. In sound-to-image generation, we take the AudioCLIP~\cite{guzhov2022audioclip} audio encoder.

\textbf{Datasets.}
We have collected the text data for GlueNet training. In monolingual T2I, only English texts are required, and we extract 18M captions from Laion-400M as training corpus, and more data will be more helpful. Multilingual GlueNets require parallel texts in different languages. These data are collected by Wikimarix~\cite{schwenk2019wikimatrix} and \href{https://pypi.org/project/googletrans/}{Googletrans} with $\sim$2M sentences for each. To train GlueNet for sound-to-image generation, we use the Urbansound8k dataset~\cite{salamon2014dataset}\footnote{https://urbansounddataset.weebly.com/urbansound8k.html}, which consists of more than 8,000 audio samples collected from city environments in ten different classes, such as dog barking, children playing, car horn, siren, etc.

 
\subsection{Monolingual Text-to-Image Generation}

Monolingual T2I generation focuses on the comparison with LDM given the English text prompts. We have decomposed our pipeline into two stages: alignment and finetuning. GlueNet helps to align the new language model whose potential can be better exploited by further finetuning of image text pairs. This section verifies the benefits of our vanilla and finetuned models in both quantitative and qualitative evaluations.  \textbf{T5+GlueNet} refers to the LDM model with the stacking of T5 and GlueNet as the text encoder where the LDM image decoder is unchanged. \textbf{T5+FT} represents finetuning of the LDM image decoder with DDPM loss based on text conditions from a fixed T5. \textbf{T5+GlueNet+FT} is the final version, including both GlueNet and joint finetuning with the image decoder. The finetuning takes $\sim$ 100 GPU days of 116M image-text pairs filtered from Laion-400M by BLIP score~\cite{li2022blip}. Unet finetuning is optional and not needed for the remaining experiments in Sec.~\ref{sec:exp-multilingual} and Sec.~\ref{sec:exp-sound2img}.

\subsubsection{Quantitative Evaluation}


Tab.~\ref{tab:fid} reported the zero-shot FID score~\cite{heusel2017gans} on COCO dataset~\cite{lin2014microsoft} based on the pyorch-fid~\cite{Seitzer2020FID} package. Our re-implementation score of LDM is slightly higher than the paper's one~\cite{rombach2022high}, which may because of different implementations. The table shows the minor drop of T5+GlueNet from LDM since our current GlueNet model still leaves some incorrect alignments, which could be fixed by either joint finetuning or collecting more text corpus to train GlueNet. With the $\sim$100 GPU-days finetuning of a subset of the Laion dataset (\ie, 116M image-text pairs), the overall performance has been considerably improved, even outperforming the LDM by certain margins. 

\begin{wraptable}{r}{5.0cm}
\small
\vspace{-3mm}
\caption{FID on COCO. * or ** indicate our implemented results with 100 or 200 GPU days. ZS means zero-shot.}
\label{tab:fid}
\begin{tabular}{lll}
\toprule[1.25pt]
\multicolumn{1}{c}{Method}  &\multicolumn{1}{c}{ FID$\downarrow$} &\multicolumn{1}{c}{ZS}
\\ 
\hline
CogView~\cite{ding2021cogview}         &27.10 &{\textcolor{red}{\XSolidBrush}} \\
LAFTTE~\cite{zhou2021lafite}             &26.94 &{\textcolor{red}{\XSolidBrush}}\\
GLIDE~\cite{nichol2021glide}             &12.24  &{\textcolor{red}{\XSolidBrush}}\\ 
Make-A-Secne~\cite{gafni2022make}             &\textbf{11.84} &{\textcolor{red}{\XSolidBrush}} \\ \cdashline{1-3}
LDM~\cite{rombach2022high}             &12.63  &{\textcolor{green}{\Checkmark}}\\ \cdashline{1-3}
LDM*             &13.55 &{\textcolor{green}{\Checkmark}} \\ 
T5+FT*          &23.30 &{\textcolor{green}{\Checkmark}}       \\
T5+FT**          &12.41 &{\textcolor{green}{\Checkmark}}       \\
T5+GlueNet             &14.32 &{\textcolor{green}{\Checkmark}} \\ 
T5+GlueNet+FT*            &\underline{12.05} &{\textcolor{green}{\Checkmark}} \\ 
\bottomrule[1.25pt]
\end{tabular}
\end{wraptable}


\begin{figure*}[t]
\centering
\includegraphics[height=0.3\linewidth,width=1\linewidth]{   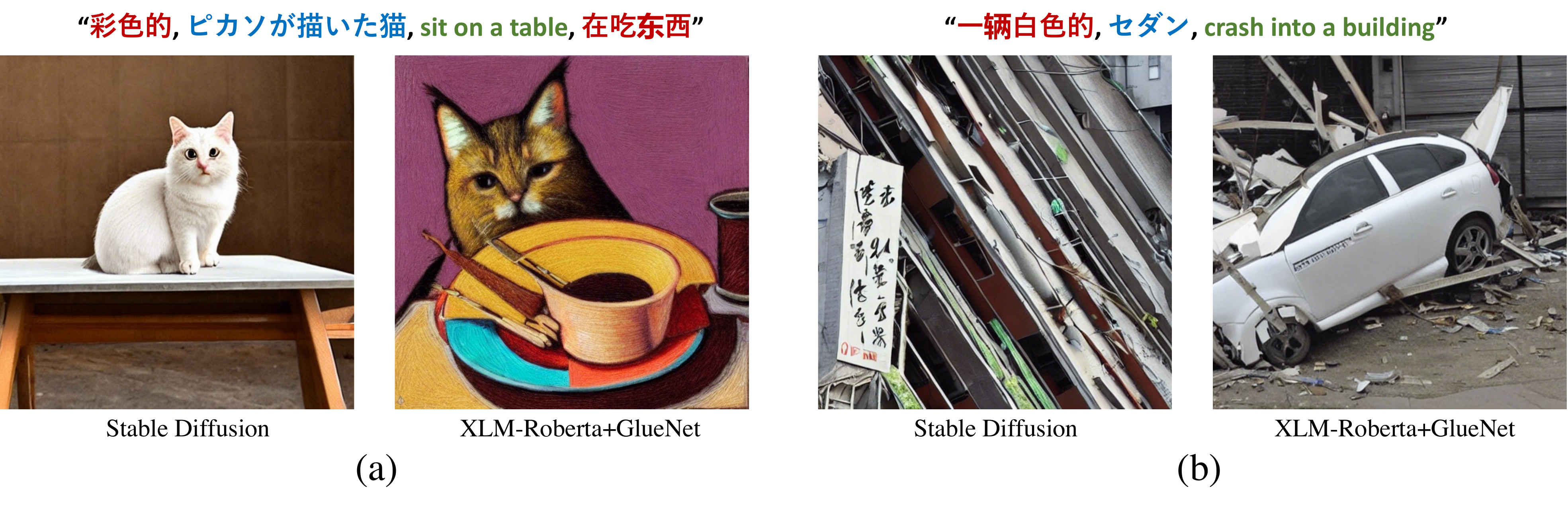}
\vspace{-6mm}
\caption{Hybrid multilingual generation in resolution of 512 $\times$ 512. There are three-different-language texts in the input caption including  \textcolor{red}{Chinese}, \textcolor{blue}{Japanese} and \textcolor[RGB]{100,220,100}{English}. The caption of (a) is ``colorful, a cat painted by Picasso, sit on a table, is eating food'' and the caption of (b) is ``a white, sedan, crash into a building''. With our GlueNet infused ahead, the XLM-Roberta can guide SDM decoder to generate reasonable results where the original SDM fails to work.}\label{fig:hybrid}
\vspace{-2mm}
\end{figure*}

\begin{figure}[t]
\centering
\includegraphics[height=0.6\linewidth,width=\linewidth]{   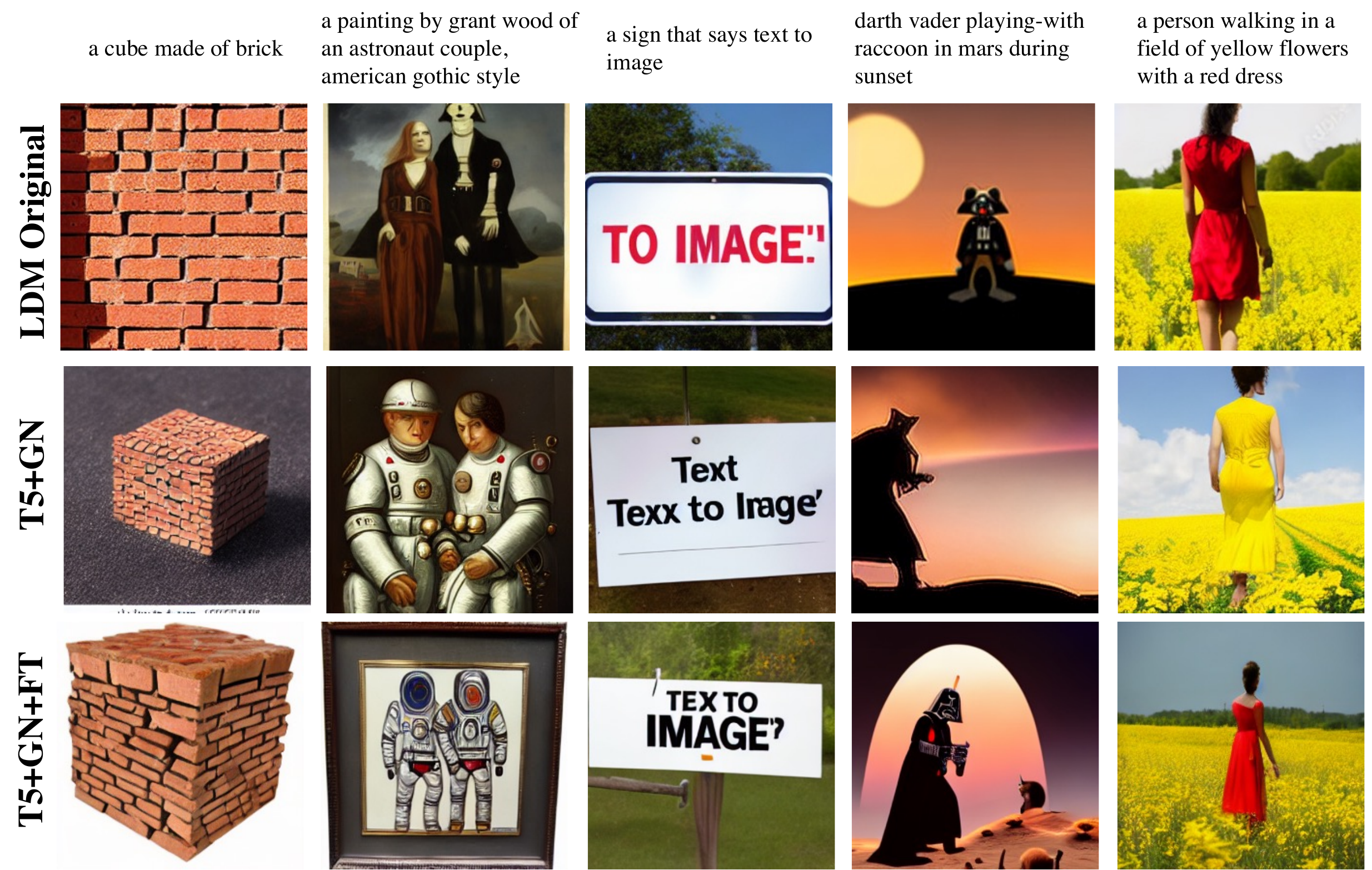}
\vspace{-5mm}
\caption{Monolingual generation in resolution 256 $\times$ 256 with guidance weight 7.5 and DDIM steps 200. The details of each method can be referred to in Sec.~\ref{sec:experiments-setup}.  }\label{fig:db_wn_demo}
\end{figure}

We have also conducted a user study to compare the text controllability in Fig.~\ref{fig:user-study} by Amazon Turk~\cite{doi:10.1177/1745691610393980}. There are $~\sim$1K prompts collected from Drawbench~\cite{saharia2022photorealistic} and Winoground~\cite{thrush2022winoground}, which comprehensively cover various scenarios and numerous difficult cases. We report the percentages of users' votes for the three methods in Fig.~\ref{fig:user-study}. 
In the user study of controllability, ours outperformed the LDM by 3\%, which is a non-trivial improvement. Moreover, the \href{https://docs.scipy.org/doc/scipy/reference/generated/scipy.stats.binomtest.html}{p-value} is computed as 0.04 with three repeats of user study that is statistically significant according to the criteria of $<$0.05.


\subsubsection{Visual Comparison}
 
In Fig.~\ref{fig:db_wn_demo}, we have exhibited some generated images of different methods. By comparing LDM and T5+GlueNet, we can conclude that the proposed GlueNet helps to achieve comparable high-quality images even replacing conditions. Moreover, the T5+GlueNet+FT shows stronger controllability in many cases compared with LDM, such as ``a cube made of brick'', whereas the LDM's results fail to display the concept of ``cube''.

\subsection{Multilingual Text-to-Image Generation}
\label{sec:exp-multilingual}
Our proposed framework is very general and generic to many language models, including the multilingual backbones. We choose the SDM as our base model with the pre-trained {clip-vit-large-patch14}\footnote{https://huggingface.co/openai/clip-vit-large-patch14} as the text encoder to be upgraded by the multilingual one, \ie, {XLM-Roberta-L}. The GlueNet helps to align such two backbones with the parallel corpus as training data. Thus, the old SDM image decoder can successfully understand the text embeddings of other languages encoded by GlueNet and XLM-Roberta-L without any additional parameters updating except GlueNet.

\subsubsection{Comparison}
\begin{table}[t]
\begin{center}
\caption{ Comparison with AudioClip model for sound-to-image generation. The CLIP$ \uparrow$ score over generated results on UrbanSound8K benchmark is reported here. We have evaluated three different settings including 1) single sound, 2) sound-sound mix and 3) sound-text mix. }
\label{tabs:sound2img}
\scalebox{0.9}{
\begin{threeparttable}
 \centering
  \begin{tabular}{c c c c}
  \toprule[1.25pt]
\multicolumn{1}{c}{ } & {Sound} &{Mixed Sounds} &{Sound-Text}
  
  \\
   \hline

{AudioCLIP+SDM} &16.07 & 13.25  & 17.22 \\

{GlueNet+SDM} &\bf 20.85 & \bf 20.50   &\bf 22.02 \\
 \bottomrule[1.25pt]
\end{tabular}
\renewcommand{\labelitemi}{}
\end{threeparttable}
}
\end{center}
\end{table}

\begin{figure*}[t]
\centering
\includegraphics[height=0.215\linewidth,width=\linewidth]{    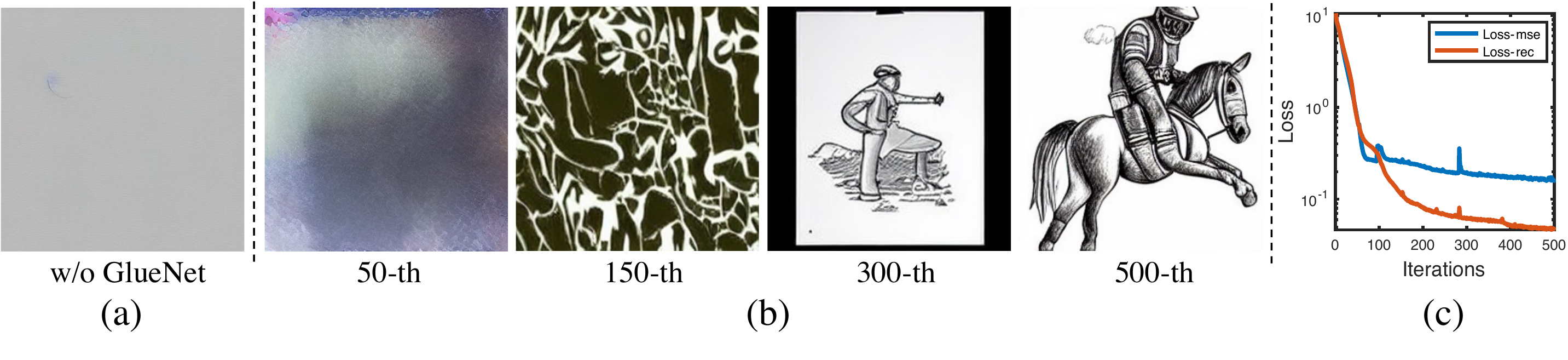}
\vspace{-8mm}
\caption{Alignment analysis. (a): Initial result of direct replacement of a new text encoder, \ie, LDM-Bert  $\rightarrow$ T5-3B. (b): Results of GlueNet in different iterations with the new text backbone. The prompt is ``an astronaut riding a horse as a pencil drawing''. (c): Curve of $\mathcal{L}_{mse}$  and $\mathcal{L}_{rec}$ in training.}\label{fig:alignment}
\vspace{-5mm}
\end{figure*}

\begin{table}[t]
\begin{center}
\caption{ \red{Comparison with translation model, \ie, M2M100-418M~\cite{fan2021beyond}, for multilingual generation. The Multilingual-CLIP$ \uparrow$ score over Crossmodal~\cite{thapliyal2022crossmodal} benchmark is reported here. GlueNetR denotes the re-weighted objective as described in Sec.~\ref{sec:cross-modal}.} }
\label{tabs:multilingual}
\scalebox{1}{
\begin{threeparttable}
 \centering
  \begin{tabular}{c c c c}
  \toprule[1.25pt]
\multicolumn{1}{c}{ } & {M2M100~\cite{fan2021beyond}} &
  
  {GlueNet} &
  
  {GlueNetR}
  
  \\
   \hline

{Chinese} &\bf 24.50 &22.01 &23.17   \\
{French} &\bf 25.08 &23.09  &23.91 \\
{Spanish} & 23.83 &22.91  &\bf24.18 \\
{Japanese} &23.73 & 23.99  &\bf24.23 \\
{Italian} &21.08 &22.40   &\bf 22.88 \\
  \hline
  {Average} &23.64 & 22.88   &\bf 23.67 \\
 \bottomrule[1.25pt]
\end{tabular}
\renewcommand{\labelitemi}{}
\end{threeparttable}
}
\end{center}
\end{table}

\begin{figure}
\centering
\includegraphics[height=0.9\linewidth,width=\linewidth]{   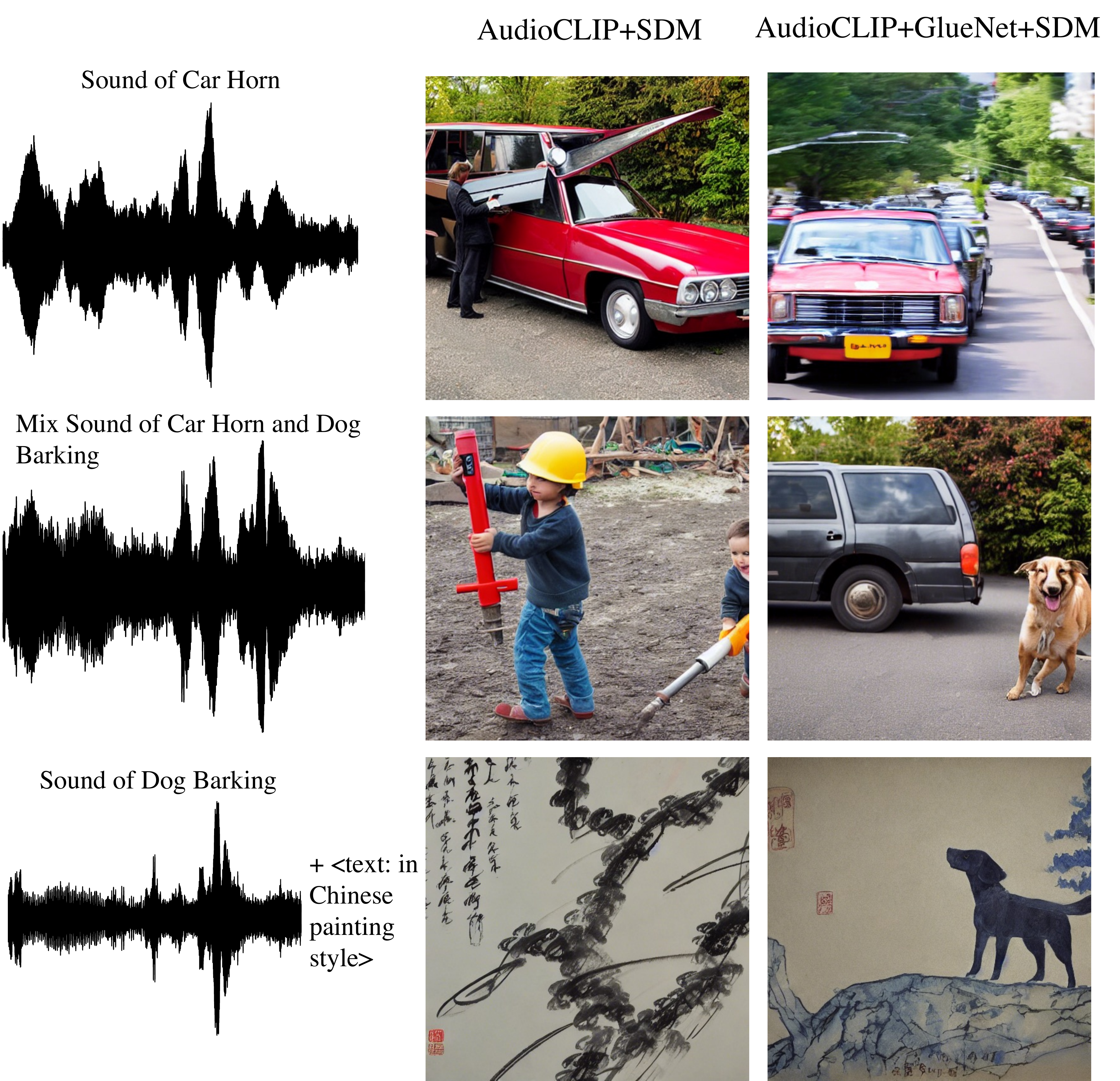}
\vspace{-5mm}
\caption{Sound-to-Image generation results.  }\label{fig:sound2img_results}
\end{figure}

\begin{figure}
	\centering
	\subfigure[Single Sound]{
		\begin{minipage}[b]{0.22\textwidth}
			\includegraphics[width=1\textwidth]{   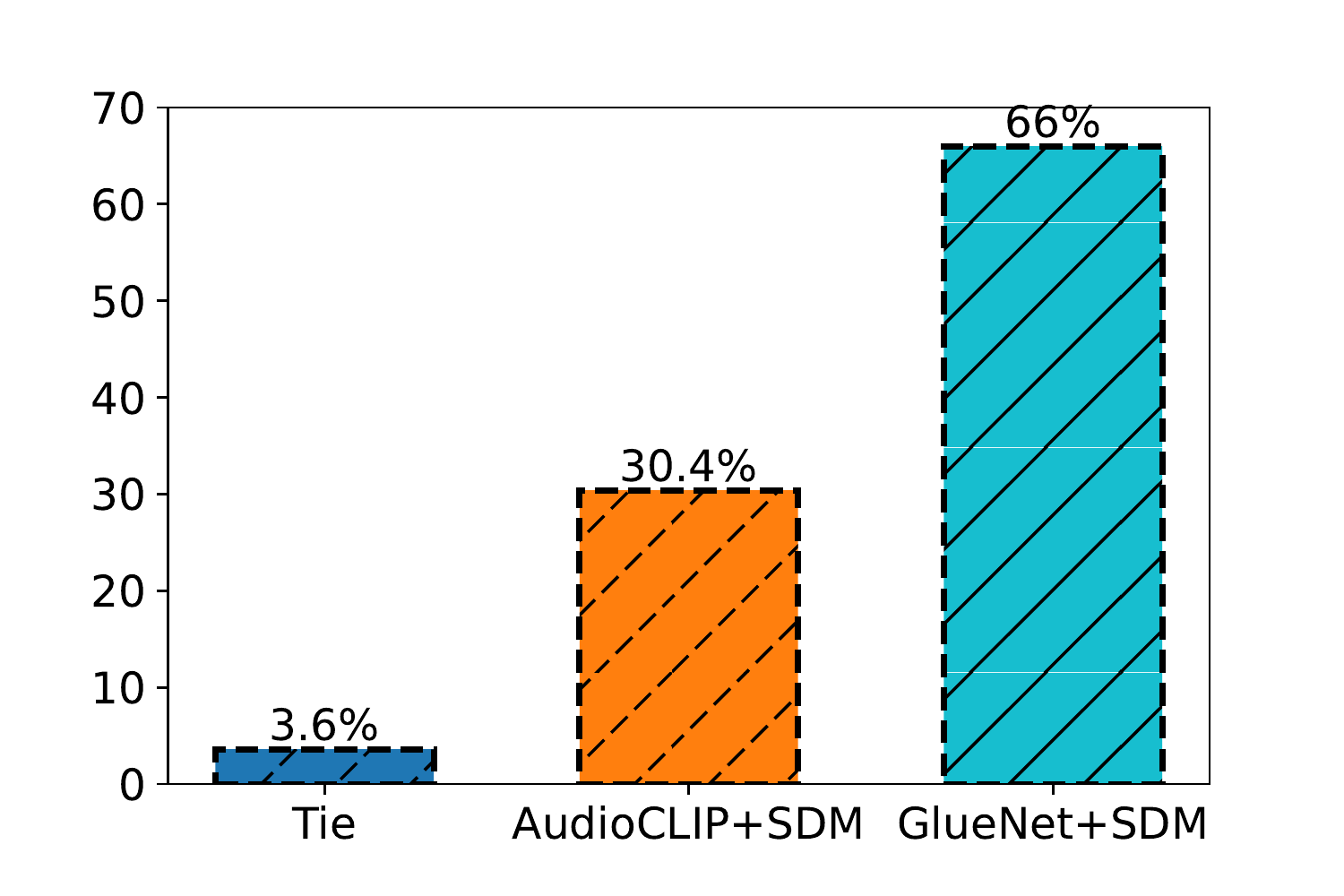}
		\end{minipage}
	
	}
    \subfigure[Mixed Sounds]{
    	\begin{minipage}[b]{0.22\textwidth}
   		\includegraphics[width=1\textwidth]{   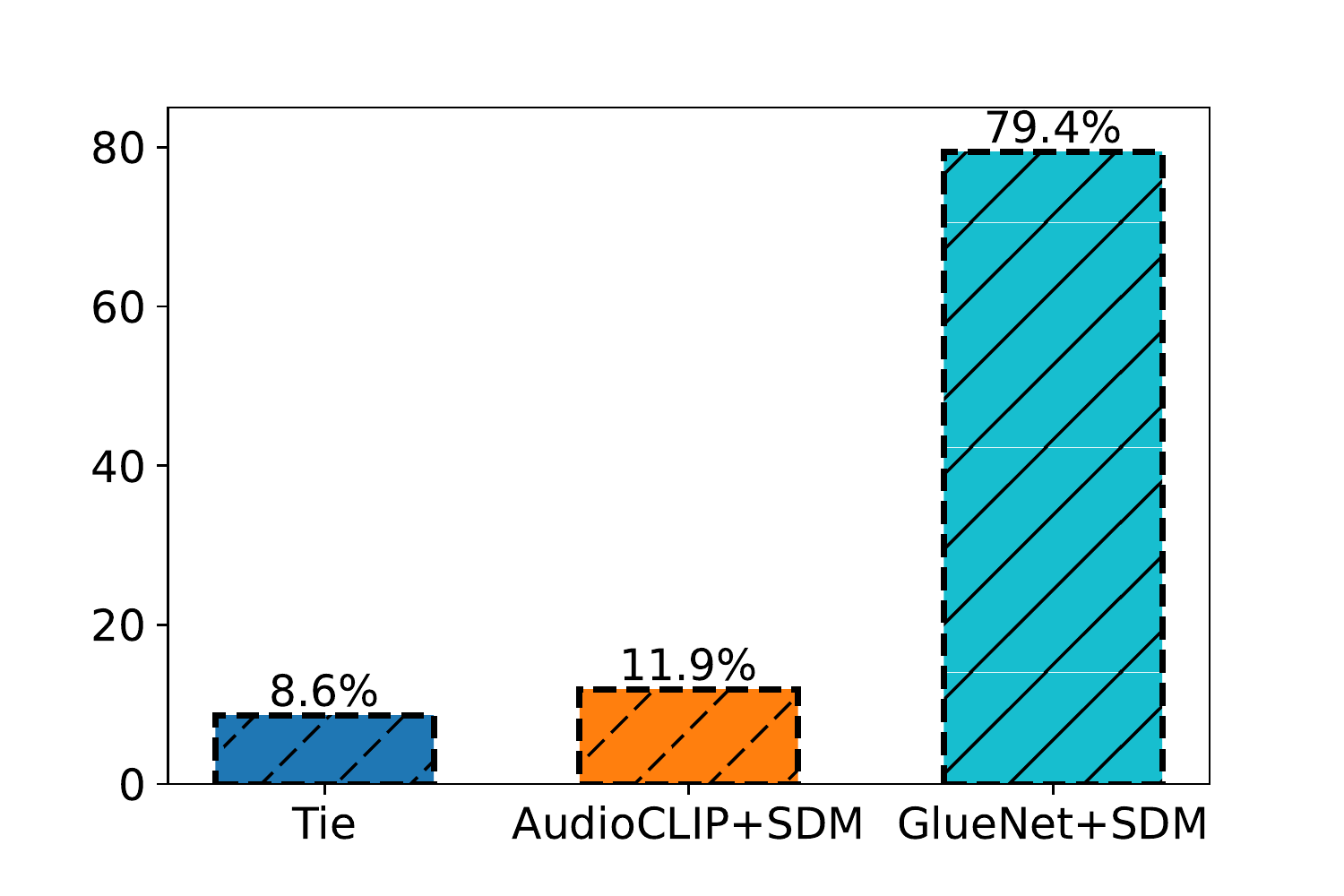}
    	\end{minipage}
    }
    \vspace{-1mm}
	\caption{Percentage of user votes of best results in sound-to-image generation with (a) single-source sound and (b) mixed-sources sound as inputs.}
	\label{fig:userstudy_sound2img}
 \vspace{-3mm}
\end{figure}

Fig.~\ref{fig:multi} shows the example of multilingual generation results with the prompt ``afternoon garden oil painting painted by impressionists'' in many languages. Most of the results reveal the content of ``afternoon garden''. The ``impressionists'' is a challenging notion that rarely appears in our training data. Thus, more high-quality parallel data will be greatly helpful. Compared with those finetuned over multilingual image text pairs, our solution is far more efficient and cheaper. For quantitative evaluation,  we have compared it with a popular multilingual translation model M2M100-418M~\cite{fan2021beyond} (418M \#params) trained by 7.5B parallel sentences. In contrast, our GlueNet uses far fewer training data (2M sentences for each pair). The results are listed in Tab.~\ref{tabs:multilingual} where we reported the Multilingual-CLIP score calculated over the 100 randomly selected multilingual image-caption pairs from Crossmodal dataset~\cite{thapliyal2022crossmodal}.The re-weighted version achieved competitive results of translation-based models under a significantly lower training cost.

\subsubsection{Hybrid Multilingual Generation}

The hybrid multilingual generation is more challenging that both the original SDM and third-party translation toolbox cannot address. Our proposed GlueNet has surprising capability in this problem. According to the results in Fig.~\ref{fig:hybrid}, the multilingual backbone bonded with GlueNet gives the precise guide to the SDM decoder even with the prompts in three different languages and randomly mixed.

\subsection{Sound-to-Image}
\label{sec:exp-sound2img}

Beyond the text signals, the proposed GlueNet also achieves sound-to-image generation by aligning the CLIP text encoder with AudioCLIP~\cite{guzhov2022audioclip} audio encoders. This is a challenging task due to the modalities gap and the mismatch of token lengths (77 for CLIP and 1 for AudioCLIP). To address these challenges, we apply a reweighted objective with a high focus on the top and informative signals, as described in Sec.~\ref{sec:cross-modal}.
The input for GlueNet is the pairing data $<sound, label>$ encoded by AudioCLIP-audio-encoder (ESResNeXt~\cite{guzhov2021esresne}) and the CLIP-text-encoder respectively. Then, GlueNet is trained to align the sound embedding with the text embedding according to the objectives described in Sec.~\ref{sec:method}. To evaluate the effectiveness of GlueNet, we conducted both quantitative and visual comparisons over the testing set of Urbansound8k. Our baseline for comparison is the deployment of AudioCLIP audio and text models to retrieve the labels with audio input. The retrieved labels are then fed into the Stable Diffusion to generate results. The CLIP score in Tab.~\ref{tabs:sound2img} strongly reveals the superiority of our GlueNet over the baseline, particularly over mixed sounds, which are more challenging than single-source sounds and can be difficult for humans to distinguish sometimes.
We also introduce cross-modal signals fusion as another contribution of this paper. Our analysis of the sound-text fusion in Tab.\ref{tabs:sound2img} and Fig.~\ref{fig:sound2img_results} demonstrate that our proposed GlueNet achieves the best results.

\subsection{Alignment Analysis}
To understand how GlueNet maps the cross-model features, we have visualized its trajectory with the generative results in different iterations in Fig.~\ref{fig:alignment}. In the beginning, the old image decoder could not understand the new text embeddings at all. Then, certain patterns such as ``pencil drawing'' appear after 300 iterations of training with a significant decrease in losses. Such a gap can finally be sealed at the end where both $\mathcal{L}_{mse}$  and $\mathcal{L}_{rec}$ converged to small values.

\subsection{Comparison with PEFT methods}
GlueNet belongs to adapter family but it is designed for offline alignment. Popular methods like ControlNet~\cite{zhang2023adding} or T2I-adapter~\cite{mou2023t2i} require new condition-image pairs to train the adapter models. 
\begin{wraptable}{r}{3.2cm}
{\small
\begin{tabular}{c|c}
\toprule
Method  & \makecell{ FID $\downarrow$ }  \\
\hline
Lora  & 16.72 \\
GlueNet &\textbf{14.32}  \\
\hdashline
Lora+FT & 14.80  \\
GlueNet+FT &\textbf{13.19}  \\
\bottomrule
\end{tabular}
}
\end{wraptable}
Our comparison is with Lora~\cite{hu2021lora}, one of the most representative Parameters-Efficient-Finetuning (PEFT) methods. We tested Lora+T5 and GlueNet+T5 for image generation on LDM (following the setting of Tab. \ref{tab:fid}, +FT indicates 10-gpu-day fine-tuning). We trained Lora+T5 to align with LDM-bert and then fine-tuned Lora+T5 with LDM-Unet. Evidently, our GlueNet outperforms both Lora and Lora+FT in quantitative terms.

\subsection{Transformer-based GlueNet}
\begin{wraptable}{r}{5cm}
{\small
\begin{tabular}{c|cc}
\toprule
Language  & \makecell{ GNMLP} &\makecell{ GNTrans} \\
\hline
French & \textbf{23.91} &23.87 \\
Spanish &24.18 &\textbf{24.55} \\
Italian &\textbf{22.88} &21.72 \\
\hdashline
Avg &\textbf{23.66} &23.38\\
\bottomrule
\end{tabular}
}
\end{wraptable}
We have implemented a transformer-based GlueNet (GlueNet-Trans) in a multilingual setting following Tab.~\ref{tabs:sound2img}. Our results indicate that GlueNet-MLP marginally outperforms GlueNet-Trans.  We intend to further investigate the self-attention models in this context.

\subsection{Point-Cloud-to-Image}
To extend the scope of this work, we have incorporated the 3D-to-image generation. We applied the point-cloud-based pre-training model, \ie, ULIP~\cite{xue2023ulip}, as the 3D condition encoder and aligned it with CLIPText using our proposed GlueNet. 
\begin{wraptable}{r}{2.8cm}
{\small
\begin{tabular}{c|c}
\toprule
Method  & \makecell{CLIP $\uparrow$ }  \\
\hline
ULIP  & 17.29 \\
GlueNet & \textbf{21.67}  \\
\bottomrule
\end{tabular}
}
\end{wraptable}
As a result, we have enabled point-cloud-based image generation by substituting the ULIPPoint encoder. 
We also compared it with the retrieval-based baseline, following the settings of Tab.~\ref{tabs:sound2img} (where sound is replaced with point cloud objects), and our method demonstrated a significant improvement.

\vspace{-2mm}
\section{Conclusion}
\vspace{-2mm}
Infusing the pre-trained conditional encoder into the existing T2I image generator is an exciting direction toward a more powerful AI system. However, the current encoder cannot be easily upgraded due to the tight match. This paper attempts to break such a strong constraint of the corresponding image-text models towards flexible modularization and efficient upgrading. To address the severe misalignment, we have proposed the GlueNet with both objectives for cross-model alignment and originality protection. Empirically, it is beneficial for the overall performance and enables versatile functionalities for X-to-image generation within a limited budget. We wish this work to be inspirable to the community in large-scale AI system design.

{\small
\bibliographystyle{ieee_fullname}
\bibliography{main}
}

\clearpage

\appendix
\onecolumn

\begin{center}
\Large
\textbf{Appendix}
\end{center}

\section{Details of GlueNet}
The key contribution of this paper is GlueNet, which addresses both cross-model alignment and feature protection. In this section, we will provide more details about GlueNet, including its implementation and configuration, an analysis of its sizes, and other ablation studies.

\subsection{Architecture and Configuration}
In Section 4.2 of the main paper, we introduced the basic architecture of GlueNet. Different sizes of GlueNet are available, and their configurations are detailed in Table~\ref{tab:configutations}. The GlueNet with three residual modules is the smaller variant, with 34M parameters in its encoder, while the GlueNet-5RMs has 51M parameters. However, larger models tend to have slower speed and higher computation costs. Moreover, there could be slight difference in the architecture for different encoders alignment. To maintain inference speed and efficiency during the finetuning stage, the size of GlueNet, when working as an injected module, should not exceed that of the image decoder and text encoder. Based on our empirical observations, assigning RM as five produces satisfactory results. GlueNet-3RMs, on the other hand, appears weak in representation learning. Besides the MLP-mixer, a self-attention-based model may also be used to implement GlueNet. In the future, we plan to explore more suitable architectures for GlueNet.

\begin{table}[h]
\caption{Model configurations for our GlueNet. We introduce two configurations of GlueNet-3RMs and GlueNet-5RMs. [LN] represent that layer-normalization is optional in the Tail Net. DIM-OUT and TOKEN-OUT are 77 and 1024 for Stable Diffusion v1. DIM-IN and TOKEN-IN depend on the target encoder to replace. $N$ is assigned as 1 if TOKEN-IN is equal to TOKEN-OUT. $N$ is larger than 1 (2 or 3) if TOKEN-IN is not equal to TOKEN-OUT.  }
\vspace{0.1cm}
\centering
\label{tab:configutations}
{\footnotesize
\resizebox{.8\textwidth}{!}{
\begin{tabular}{c|c|c|c|c}
\Xhline{3\arrayrulewidth}
 Stage&Dimensions & Block & GlueNet-3RMs & GlueNet-5RMs  \\
 \hline
\makecell{ \\ \\ Head Net} & TOKEN-IN $\rightarrow$ TOKEN-OUT & 
\makecell{Token\\MLP} & $\left[\makecell{\mathrm{Linear, LN, \sigma} \\ \mathrm{Linear, LN, \sigma} \\\mathrm{Linear, LN}} \right] \times N $&  
$\left[\makecell{\mathrm{Linear, LN, \sigma} \\ \mathrm{Linear, LN, \sigma} \\\mathrm{Linear, LN}} \right] \times N$

\\
\cline{2-5}
 \rule{0pt}{5ex} & DIM-IN $\rightarrow$DIM-OUT & \makecell{Sequence\\MLP} 
 &  $\left[\makecell{\mathrm{Linear, LN, \sigma} \\ \mathrm{Linear, LN, \sigma} \\\mathrm{Linear, LN}} \right] \times N$
 &  $\left[\makecell{\mathrm{Linear, LN, \sigma} \\ \mathrm{Linear, LN, \sigma} \\\mathrm{Linear, LN}} \right] \times N$
 \\

%
\hline
\makecell{ \\ \\ Body Net} & TOKEN-OUT $\rightarrow$ TOKEN-OUT  & 
\makecell{Token\\MLP} & $\left[\makecell{\mathrm{Linear, LN, \sigma} \\ \mathrm{Linear, LN, \sigma} \\\mathrm{Linear, LN}} \right] \times 3$&  
$\left[\makecell{\mathrm{Linear, LN, \sigma} \\ \mathrm{Linear, LN, \sigma} \\\mathrm{Linear, LN}} \right] \times 5$
\\
\cline{2-5}
 \rule{0pt}{5ex} &DIM-OUT $\rightarrow$DIM-OUT & \makecell{Sequence\\MLP} 
 &  $\left[\makecell{\mathrm{Linear, LN, \sigma} \\ \mathrm{Linear, LN, \sigma} \\\mathrm{Linear, LN}} \right]\times 3$
 &  $\left[\makecell{\mathrm{Linear, LN, \sigma} \\ \mathrm{Linear, LN, \sigma} \\\mathrm{Linear, LN}} \right]\times 5$
 \\
%
\hline
\makecell{ \\ \\ Tail Net} & TOKEN-OUT $\rightarrow$ TOKEN-OUT & 
\makecell{Token\\MLP} & $\left[\makecell{\mathrm{Linear,[LN], \sigma} \\ \mathrm{Linear, [LN], \sigma} \\\mathrm{Linear, [LN]}} \right]$&  
$\left[\makecell{\mathrm{Linear, [LN], \sigma} \\ \mathrm{Linear, [LN], \sigma} \\\mathrm{Linear, [LN]}} \right]$
\\
\cline{2-5}
 \rule{0pt}{5ex} &DIM-OUT $\rightarrow$DIM-OUT & \makecell{Sequence\\MLP} 
 &  $\left[\makecell{\mathrm{Linear, [LN], \sigma} \\ \mathrm{Linear, [LN], \sigma} \\\mathrm{Linear, [LN]}} \right]$
 &  $\left[\makecell{\mathrm{Linear, [LN], \sigma} \\ \mathrm{Linear, [LN], \sigma} \\\mathrm{Linear, [LN]}} \right]$
 \\

\Xhline{3\arrayrulewidth}
\end{tabular}
}
}
\end{table}

\section{Analysis of Text Encoder Replacement}
\subsection{Analysis of GlueNet Sizes}

\begin{figure}[t]
\centering
\includegraphics[height=0.8\linewidth,width=0.7\linewidth]{  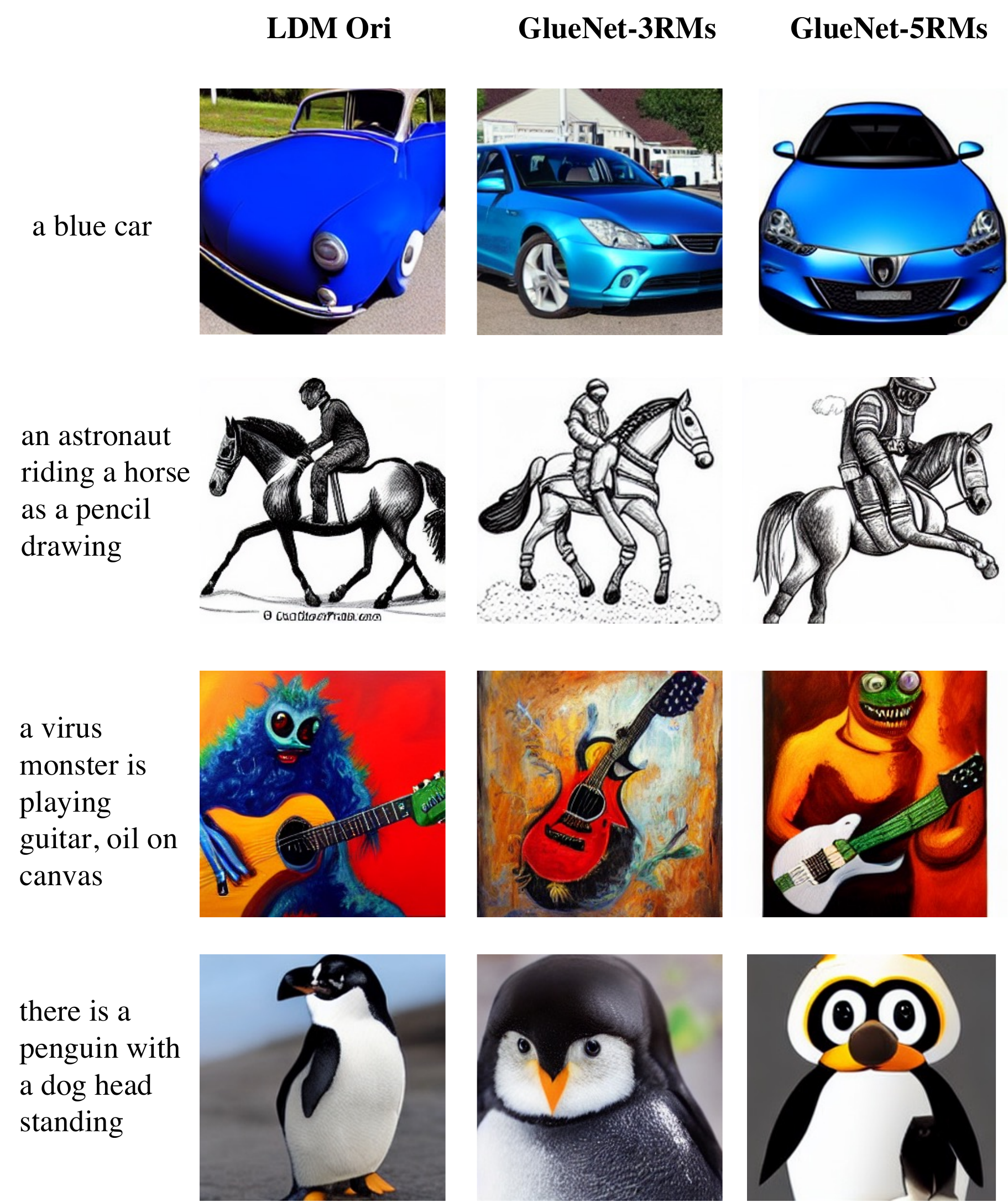}
\caption{Monolingual generation (T5 + GlueNet + LDMUnet) of example prompts in 256 $\times$ 256 with guidance weight 7.5 and DDIM steps 200.  Both T5 and LDMUnet are pre-trained ones. We only train GlueNet with different model sizes to fulfill.}\label{fig:rm_sup}
\end{figure}

Figure~\ref{fig:rm_sup} presents a visual comparison of example prompts across three different methods. The original LDM model with the checkpoint~\footnote{https://github.com/CompVis/latent-diffusion} shared in its repository is referred to as LDM Ori. Our models, GlueNet-3RMs and GlueNet-5RMs, contain three or five residual modules within their body nets, respectively. The text encoder is replaced with T5-3B, while the image decoder is the same as LDM Ori. Both GlueNet models are trained on the same text corpus, consisting of 18 million English sentences. The figure shows that GlueNet-3RMs struggles with some complex prompts, such as "a virus monster is playing guitar, oil on canvas" and "there is a penguin with a dog head standing," where GlueNet-5RMs performs better. We can, therefore, conclude that deep GlueNet is necessary for precise alignment. GlueNet-5RMs is the default configuration mostly.

\subsection{Ablation of Losses}
\begin{figure}[t]
\centering
\includegraphics[height=0.3\linewidth,width=0.8\linewidth]{  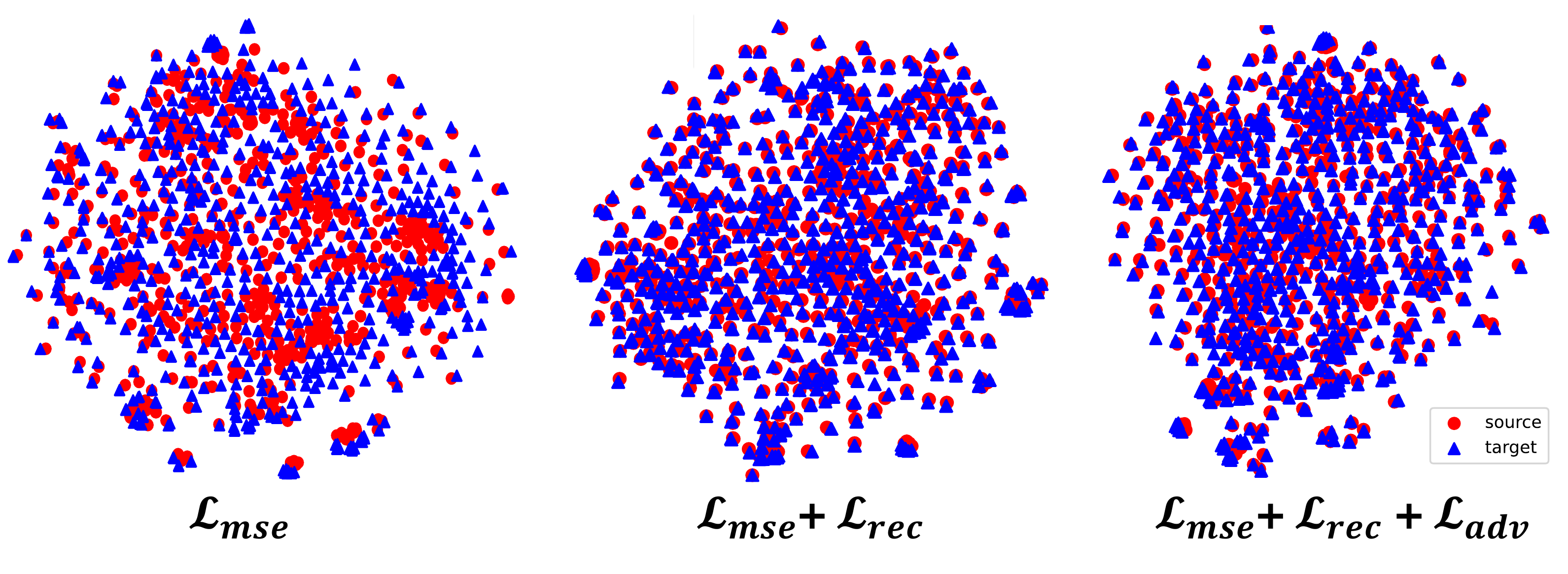}
\caption{t-SNE~\cite{van2008visualizing} cross-model feature map visualization. }\label{fig:tsne_sup}
\end{figure}

In order to analyze the impacts of different losses, we present their t-SNE map visualizations in Figure~\ref{fig:tsne_sup}. It is evident from the figure that using just the MSE loss is insufficient to align the features of the two models accurately. The reconstruction loss is crucial to maintain discrimination and avoid overfitting. In contrast, the adversarial loss does not seem to provide significant improvement in the figure. Through empirical study, we found that the adversarial loss only yields limited gains.

We conducted a quantitative ablation study on a randomly selected subset of 5,000 images from COCO dataset. For alignment, we used T5-3B as the text encoder and aligned it with LDM Unet using our GlueNet. The finetuning of Unet is denoted by FT (Unet finetuning is only needed here). We report FID and CLIP scores in the following table for comparing image quality and image-text alignment. The testing data is inferred by DDIM with 200 steps, and the image size is 256 $\times$ 256. For a comprehensive analysis, the experiments were performed using three classifier-free guidances with $s$=1.5, 5, and 7.5, according to Eq. (\ref{eq:sampling}). Table ~\ref{tabs:ablation_loss} summarizes the results of our ablation study.

The first row of the table represents the direct combination of T5 and LDM, which yields nonsensical results due to severe misalignment. The second row reports results of T5+LDM trained from scratch. The finetuning of Unet in T5+LDM took 100 GPU days, whereas GlueNet training required only 5 GPU days. However, even with ten times the cost, its results were inferior to those of GlueNets'. By comparing the third, fourth, fifth, and sixth rows, we can easily conclude the superiority of the full-version model.

\begin{table*}[t]
\begin{center}
\caption{ {Ablation study of GlueNet (T5+GlueNet+LDMUnet) over the 5K COCO subset. Finetuning is not applied in most of the cases. }}
\label{tabs:ablation_loss}
\scalebox{0.9}{
\begin{threeparttable}
 \centering
  \begin{tabular}{cccc cc cc cc}
  \toprule[1.25pt]
  \multicolumn{4}{c}{Ablations} & \multicolumn{2}{c}{$s$=1.5} & \multicolumn{2}{c}{$s$=5 } & \multicolumn{2}{c}{$s$=7.5 }  \\
    \cmidrule(lr){5-6}   \cmidrule(lr){7-8} \cmidrule(lr){9-10}
 {loss-mse} &{loss-rec} &{loss-adv} &{Finetuning} &{CLIP$\uparrow$} &{FID$\downarrow$} 
  &{CLIP$\uparrow$} &{FID$\downarrow$} &{CLIP$\uparrow$} &{FID$\downarrow$} 
  \\
   \hline
   
{\XSolidBrush} &{\XSolidBrush} &{\XSolidBrush} &\multicolumn{1}{c}{\XSolidBrush} &10.70 &98.17 &10.75 &141.19 &{11.00} &\multicolumn{1}{c}{156.70}  \\

{\XSolidBrush} &{\XSolidBrush} &{\XSolidBrush} &\multicolumn{1}{c}{\Checkmark} &18.98 &35.76 &22.05 &49.14 &{22.49} &\multicolumn{1}{c}{53.11}  \\

{\Checkmark} &{\XSolidBrush} &{\XSolidBrush} &\multicolumn{1}{c}{\XSolidBrush} &20.40 &34.41 &23.01 &48.67 &{23.40} &\multicolumn{1}{c}{52.78}  \\

{\Checkmark} &{\Checkmark} &{\XSolidBrush} &\multicolumn{1}{c}{\XSolidBrush} &20.53 &33.14 &23.23 &45.96 &{23.57} &\multicolumn{1}{c}{48.67}  \\

{\Checkmark} &{\Checkmark} &{\Checkmark} &\multicolumn{1}{c}{\XSolidBrush} &20.67 &32.80 &23.24 &45.48 &{23.74} &\multicolumn{1}{c}{48.51}  \\

{\Checkmark} &{\Checkmark} &{\Checkmark} &\multicolumn{1}{c}{\Checkmark} &\bf{21.14} &\bf{30.93} &\bf{23.88} &\bf{41.92} &\bf{24.17} &\multicolumn{1}{c}{\bf{44.58}}  \\



   
 \bottomrule[1.25pt]
\end{tabular}
\renewcommand{\labelitemi}{}
\end{threeparttable}
}
\end{center}
\end{table*}

\subsection{ {Text Encoders Analysis}}

Our proposed framework is compatible with a wide range of text encoders. In this subsection, we experimented with several language models as the text encoder, including Bert-base (110M parameters), Roberta-Large (355M parameters), Clip-text (123M parameters), T5-large (770M parameters), and T5-3B (2.8B parameters). These pre-trained text encoders were aligned with the Latent Diffusion Model (LDM) using our proposed model without requiring any finetuning of the LDM UNet. These plug-and-play models were evaluated on a subset of COCO consisting of 5,000 randomly selected samples. The FID and CLIP scores are reported in the table below for image quality and image-text alignment comparison. The GlueNets were trained on 18 million sentences sampled from the captions of Laion-400M. The testing data were inferred using DDIM with 200 steps, and the image size was set to 256×256. To provide a comprehensive analysis, we conducted experiments on three classifier-free guidance settings with $s$=1.5, 5 and 7.5. The results are summarized in Table~\ref{tabs:ablation_textmodels}. As shown in the table, we observe that both FID and CLIP scores increase with the use of larger text models, which is consistent with the findings reported in Imagen~\cite{saharia2022photorealistic}. However, CLIP-text does not perform well in our experiments due to a larger domain gap with text-only models like Bert.

\begin{table*}[t]
\small
\begin{center}
\caption{ {Analysis of Different Text Encoders over the 5K COCO subset.}}
\label{tabs:ablation_textmodels}
\scalebox{0.975}{
\begin{threeparttable}
 \centering
  \begin{tabular}{c  cc cc  cc cc cc}
  \toprule[1.25pt]
  { } & \multicolumn{2}{c}{Bert-B (110M)}& \multicolumn{2}{c}{Roberta-L (355M)} & \multicolumn{2}{c}{CLIP-Text (123M)}  & \multicolumn{2}{c}{T5-L (770M)}  & \multicolumn{2}{c}{T5-3B (2.8B)} \\
 \cmidrule(lr){2-3}  \cmidrule(lr){4-5}    \cmidrule(lr){6-7}   \cmidrule(lr){8-9} \cmidrule(lr){10-11}
 {}  &{CLIP$\uparrow$} &{FID$\downarrow$} 
  &{CLIP$\uparrow$} &{FID$\downarrow$} &{CLIP$\uparrow$} &{FID$\downarrow$} &{CLIP$\uparrow$} &{FID$\downarrow$}  &{CLIP$\uparrow$} &{FID$\downarrow$} 
  \\
   \hline

\multicolumn{1}{c}{$s$=1.5}  &19.42 &37.46  &20.03 &34.06 &19.02 &38.93 &19.78 &35.17 &{\bf 20.67} &\multicolumn{1}{c}{\bf 32.80}  \\

\multicolumn{1}{c}{$s$=5}  &21.97 &55.35   &22.85 &45.88 &21.85 &47.32 &22.76 &47.91 &{\bf 23.24} &\multicolumn{1}{c}{\bf 45.48}  \\

\multicolumn{1}{c}{$s$=7.5}  &22.68 &55.21   &23.25 &48.96 &22.38 &49.66 &23.23 &50.02 &{\bf 23.74} &\multicolumn{1}{c}{\bf 48.51}  \\


   
 \bottomrule[1.25pt]
\end{tabular}
\renewcommand{\labelitemi}{}
\end{threeparttable}
}
\end{center}
\end{table*}

\subsection{ {Variant Token Lengths}}

GlueNet demonstrates strong ability in handling text of different lengths. The token number is fixed at 77 for both LDM and SDM. Our proposed GlueNet can handle variable length text encoders without any finetuning of the Unet model. To verify this, we conducted an experimental study, and the results are reported in Table~\ref{tabs:ablation_tokenlength}. In this experiment, we used Roberta-L and its tokenizers to encode text with maximum tokens of 77, 128, and 256, respectively. The guidance setting was 5, and other experimental settings were consistent with those in Table~\ref{tabs:ablation_textmodels}.

\begin{table*}[t]
\begin{center}
\caption{{Model Transfer with variant token lengths (SrcTokenLength $\rightarrow$TargetTokenLength) over the 5K COCO subset with the guidance as 5. Roberta-L~\cite{liu2019roberta} is applied as the new text encoder to replace LDM text encoder.  } }
\label{tabs:ablation_tokenlength}
\scalebox{1}{
\begin{threeparttable}
 \centering
  \begin{tabular}{c c c c}
  \toprule[1.25pt]

  \multicolumn{1}{c}{ } & \multicolumn{1}{c}{77$\rightarrow$77 } & \multicolumn{1}{c}{128$\rightarrow$77}   & \multicolumn{1}{c}{256$\rightarrow$77}  
  \\
   \hline

\multicolumn{1}{c}{CLIP$ \uparrow$} &22.85 &23.19 &\multicolumn{1}{c}{\bf 23.37}  \\

\multicolumn{1}{c}{ FID$\downarrow$} &45.88 &45.67  &\multicolumn{1}{c}{\bf 45.53}    \\


   
 \bottomrule[1.25pt]
\end{tabular}
\renewcommand{\labelitemi}{}
\end{threeparttable}
}
\end{center}
\end{table*}

\subsection{ {Data Sizes}}

Conditional generation is a challenging task, and the alignment between the text encoder and image Unet remains an open question. We have found empirically that replacing the text encoder can yield comparable results to the source model, but it requires significant effort to surpass it. The bottleneck may lie in the Unet architecture. In this section, we provide precise computations costs for different training set sizes in Table~\ref{tabs:gluenet_cost_data}. The benchmark was performed on COCO-5K using T5-3B to replace the LDM text encoder with GlueNet, consistent with previous experiments. As shown in the table, we observe that performance increases with both the size of the training set and the computation cost.

\begin{table*}[t]
\begin{center}
\caption{{Analysis of GlueNet's training cost (T5-3B $\rightarrow$ LDM) with increasing sizes of training data. The FID and CLIP scores are computed over 5K subset of COCO. }}
\label{tabs:gluenet_cost_data}
\scalebox{1}{
\begin{threeparttable}
 \centering
  \begin{tabular}{c c c c}
  \toprule[1.25pt]
{}  & \multicolumn{3}{c}{Data Size}\\
  \multicolumn{1}{c}{ } & \multicolumn{1}{c}{5M} & \multicolumn{1}{c}{18M}   & \multicolumn{1}{c}{116M}  
  \\
   \hline

\multicolumn{1}{c}{CLIP$ \uparrow$} &20.92 &23.24 &\multicolumn{1}{c}{\bf 23.71}  \\

\multicolumn{1}{c}{FID$\downarrow$} &48.63 &45.48  &\multicolumn{1}{c}{\bf 43.17}    \\

\multicolumn{1}{c}{GPU Days$\downarrow$ } &\bf 1.67 &5.89 &\multicolumn{1}{c}{41.20}    \\


   
 \bottomrule[1.25pt]
\end{tabular}
\renewcommand{\labelitemi}{}
\end{threeparttable}
}
\end{center}
\end{table*}

\section{Monolingual (English) Text-to-Image Generation}
\begin{figure}[t]
\centering
\includegraphics[height=0.9\linewidth,width=0.8\linewidth]{  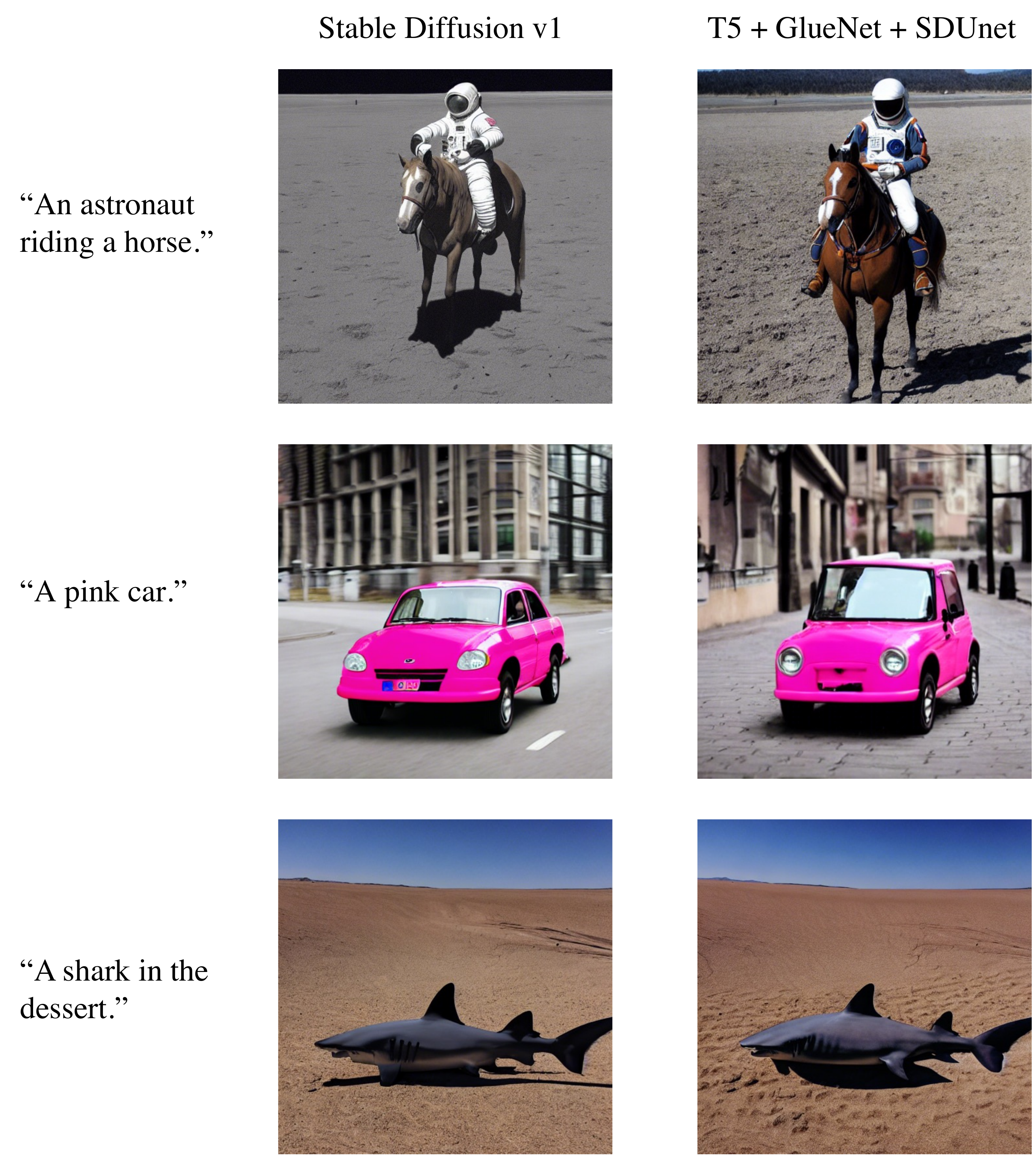}
\caption{Monolingual generation of example prompts in 512 $\times$ 512 with guidance weight 7.5 and 50 PLMS~\cite{liu2022pseudo} sampling steps.  }\label{fig:sdm_sup}
\end{figure}

To demonstrate the generality of our proposed framework, we also replaced the CLIP text encoder of Stable Diffusion (v1-4) with T5-Large. As shown in Figure~\ref{fig:sdm_sup}, our model exhibits precise controllability and excellent visual quality compared to the standard Stable Diffusion model. However, it falls short of outperforming the original Stable Diffusion model due to minor mismatches.

\clearpage

\section{Multilingual Text-to-Image Generation}
\begin{figure}[t]
\centering
\includegraphics[height=0.78\linewidth,width=0.9\linewidth]{  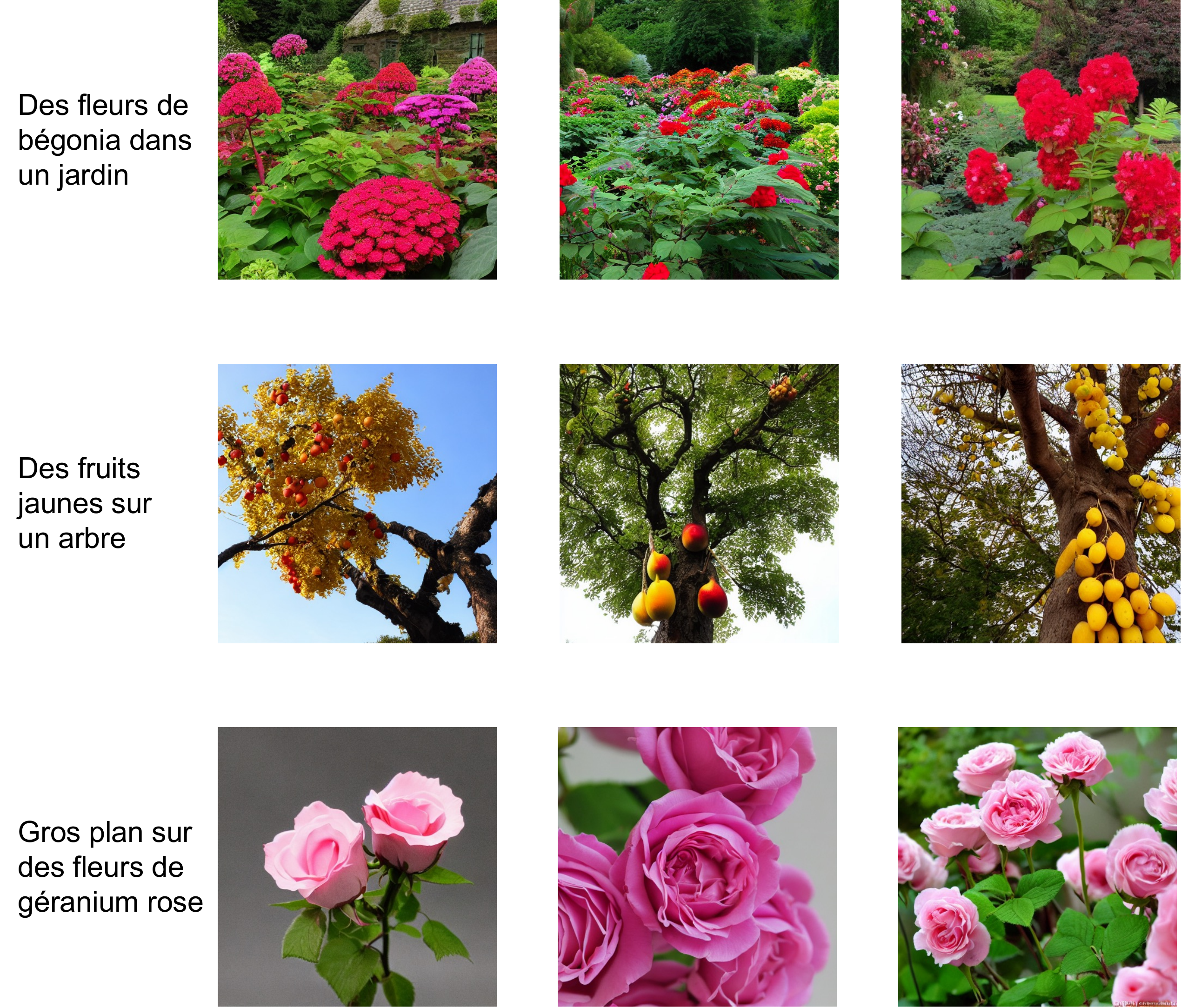}
\caption{Multilingual generation results in 512 $\times$ 512  of XLM-Roberta + GlueNet + Stable Diffusion Unet (v1-5) with the French captions. The three results are generated with different random noises.   }\label{fig:multi_sup_fr}
\end{figure}

\begin{figure}[t]
\centering
\includegraphics[height=0.78\linewidth,width=0.9\linewidth]{  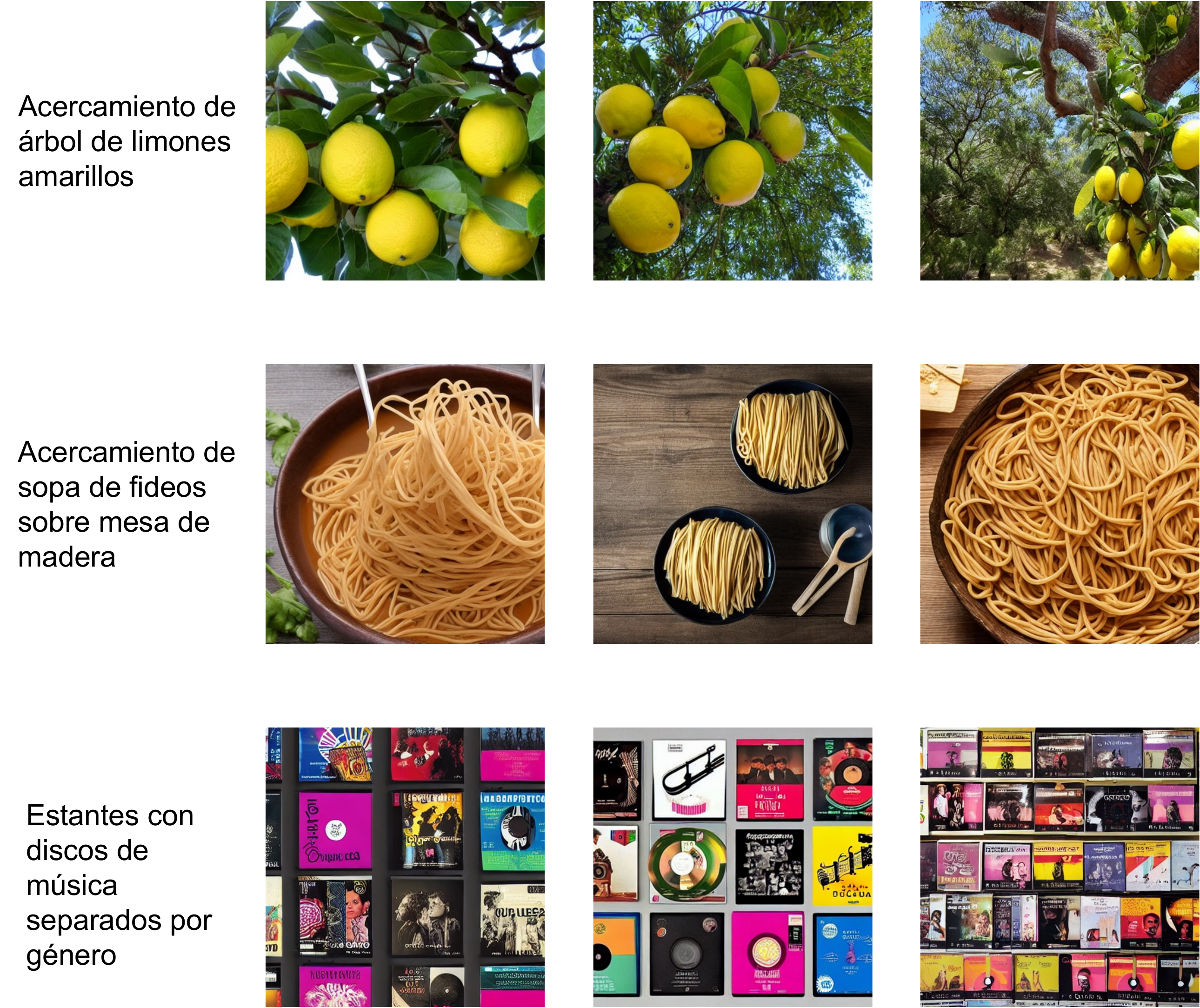}
\caption{Multilingual generation results in 512 $\times$ 512  of XLM-Roberta + GlueNet + Stable Diffusion Unet (v1-5) with the Spanish captions. The three results are generated with different random noises.   }\label{fig:multi_sup_es}
\end{figure}

\begin{figure}[t]
\centering
\includegraphics[height=0.78\linewidth,width=0.9\linewidth]{  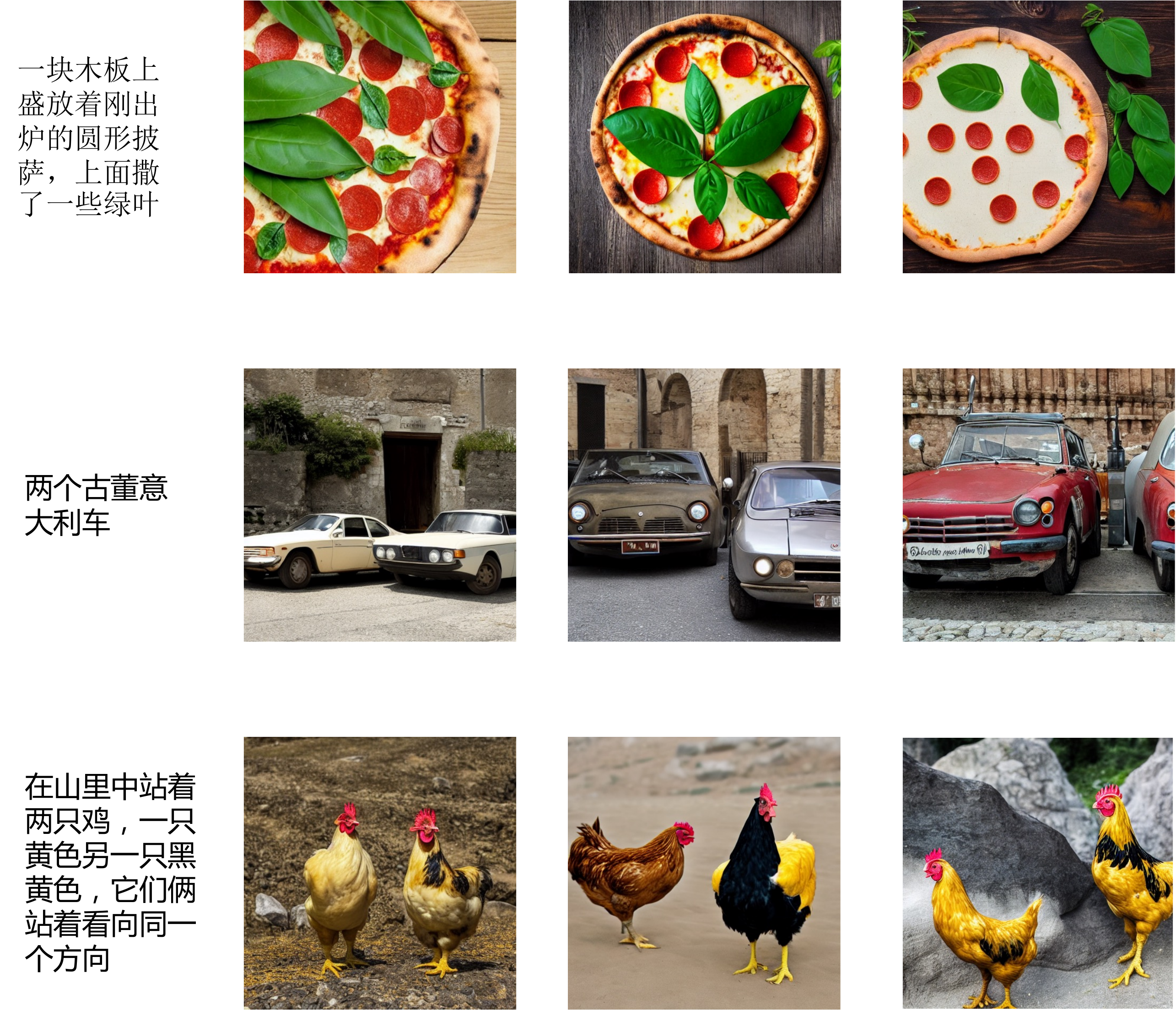}
\caption{Multilingual generation results in 512 $\times$ 512  of XLM-Roberta + GlueNet + Stable Diffusion Unet (v1-5) with the Chinese captions. The three results are generated with different random noises.   }\label{fig:multi_sup_zh}
\end{figure}

\begin{figure}[t]
\centering
\includegraphics[height=0.78\linewidth,width=0.9\linewidth]{  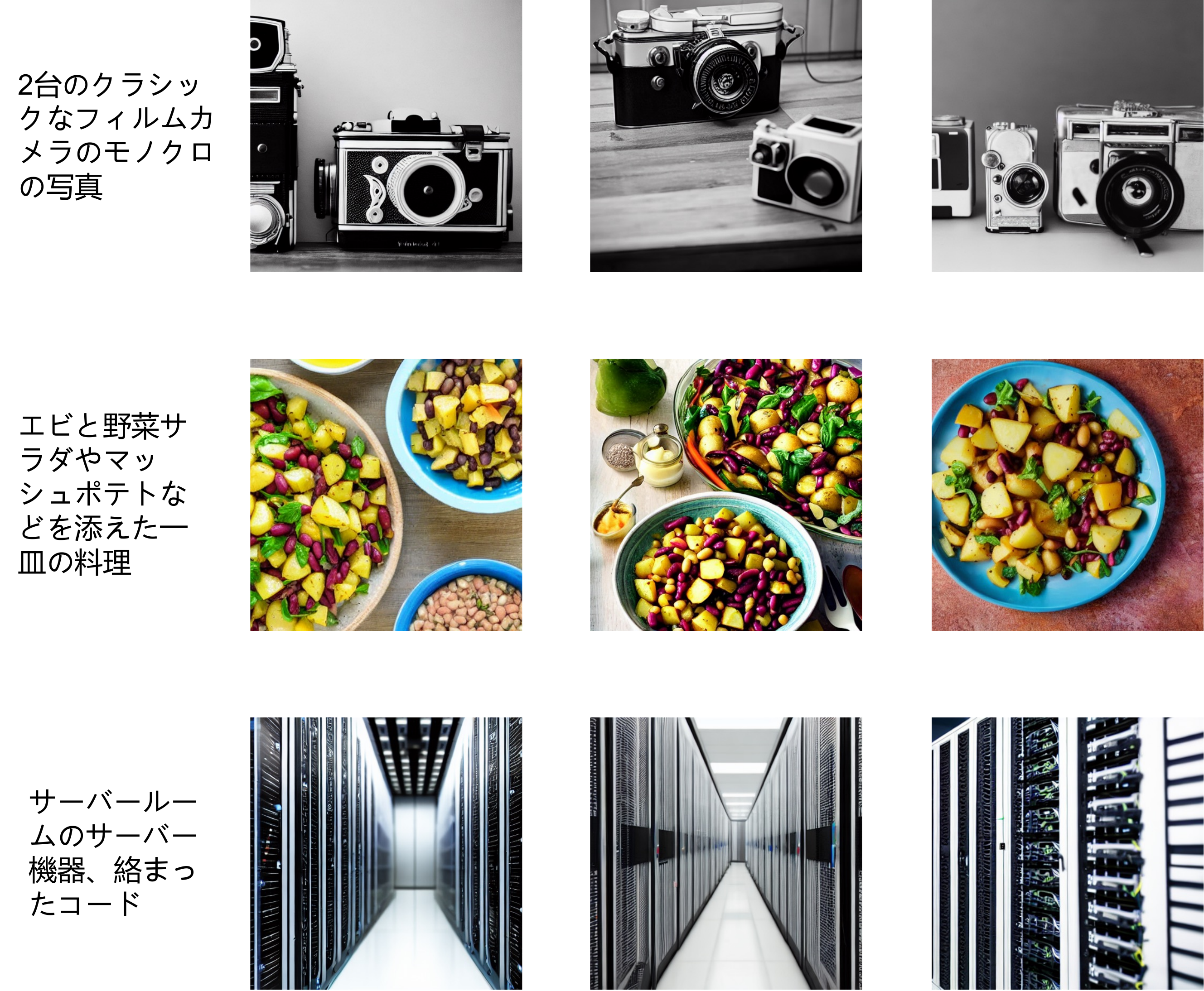}
\caption{Multilingual generation results in 512 $\times$ 512  of XLM-Roberta + GlueNet + Stable Diffusion Unet (v1-5) with the Japanese captions. The three results are generated with different random noises.   }\label{fig:multi_sup_ja}
\end{figure}

\begin{figure}[t]
\centering
\includegraphics[height=0.78\linewidth,width=0.9\linewidth]{  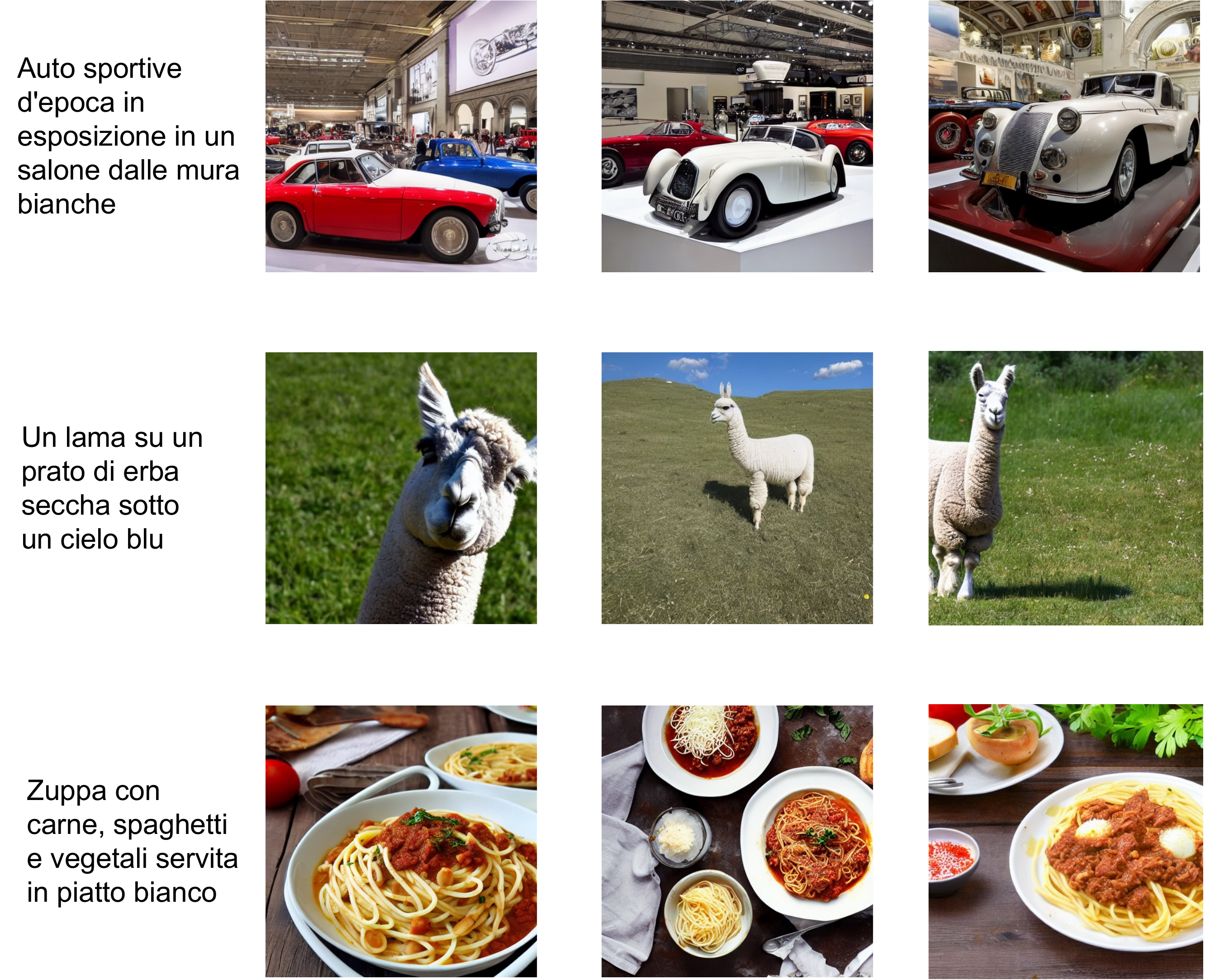}
\caption{Multilingual generation results in 512 $\times$ 512  of XLM-Roberta + GlueNet + Stable Diffusion Unet (v1-5) with the Italian captions. The three results are generated with different random noises.   }\label{fig:multi_sup_it}
\end{figure}

The additional results in Figures~\ref{fig:multi_sup_fr}, ~\ref{fig:multi_sup_es},~\ref{fig:multi_sup_zh},~\ref{fig:multi_sup_ja} and~\ref{fig:multi_sup_it} help to verify the multilingual text-to-image generation capabilities of our proposed framework. Each language is associated with a dedicated GlueNet. To build an automated pipeline, the model should be able to select the appropriate GlueNet based on the detected language.

\section{Sound-to-Image Generation}
\begin{figure}[t]
\centering
\includegraphics[height=0.9\linewidth,width=0.8\linewidth]{  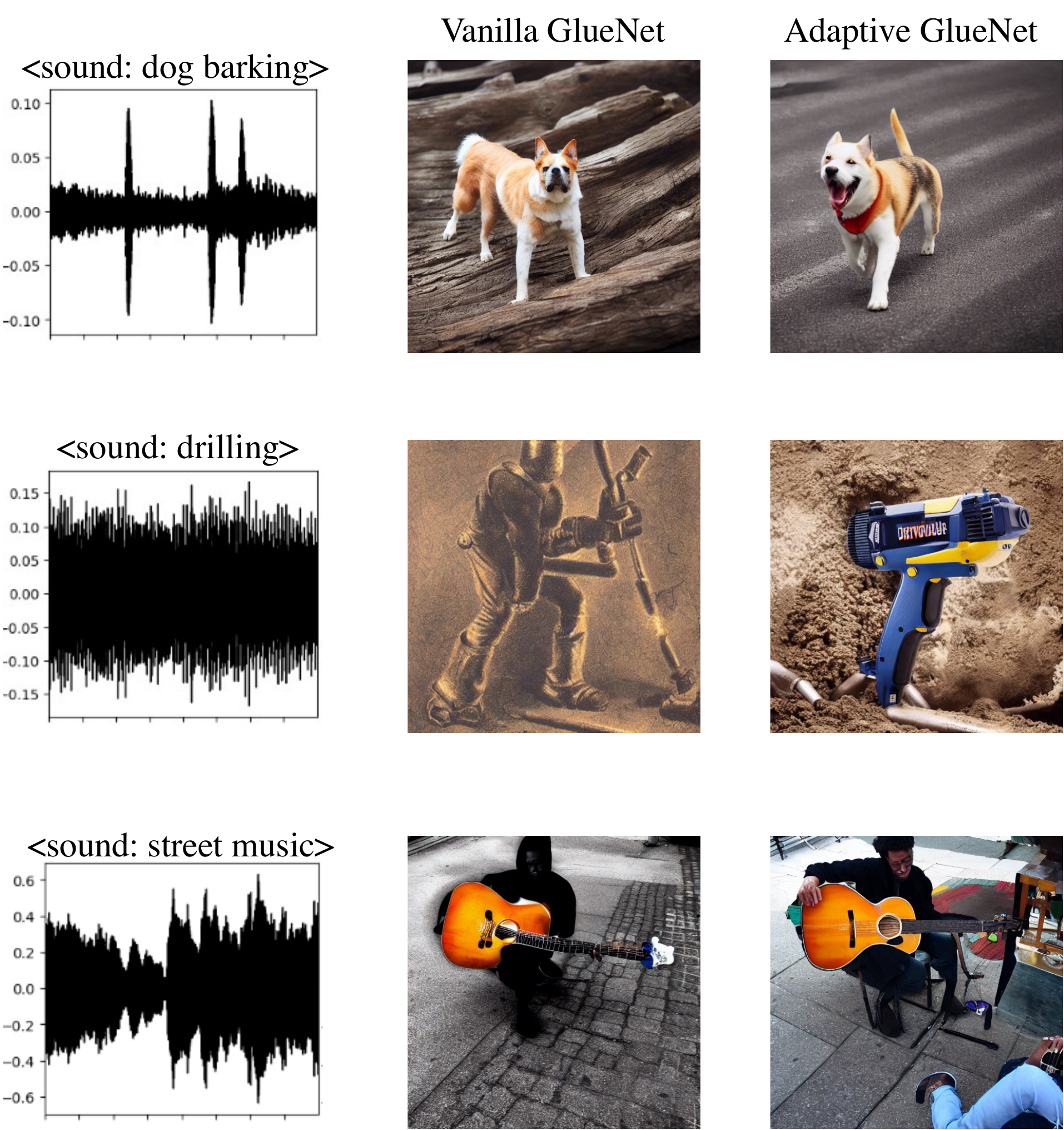}
\caption{Sound-to-image generation (Part 1/2) with vanilla and adaptive GlueNets.   }\label{fig:sound_to_image_1}
\end{figure}

\begin{figure}[t]
\centering
\includegraphics[height=0.9\linewidth,width=0.8\linewidth]{  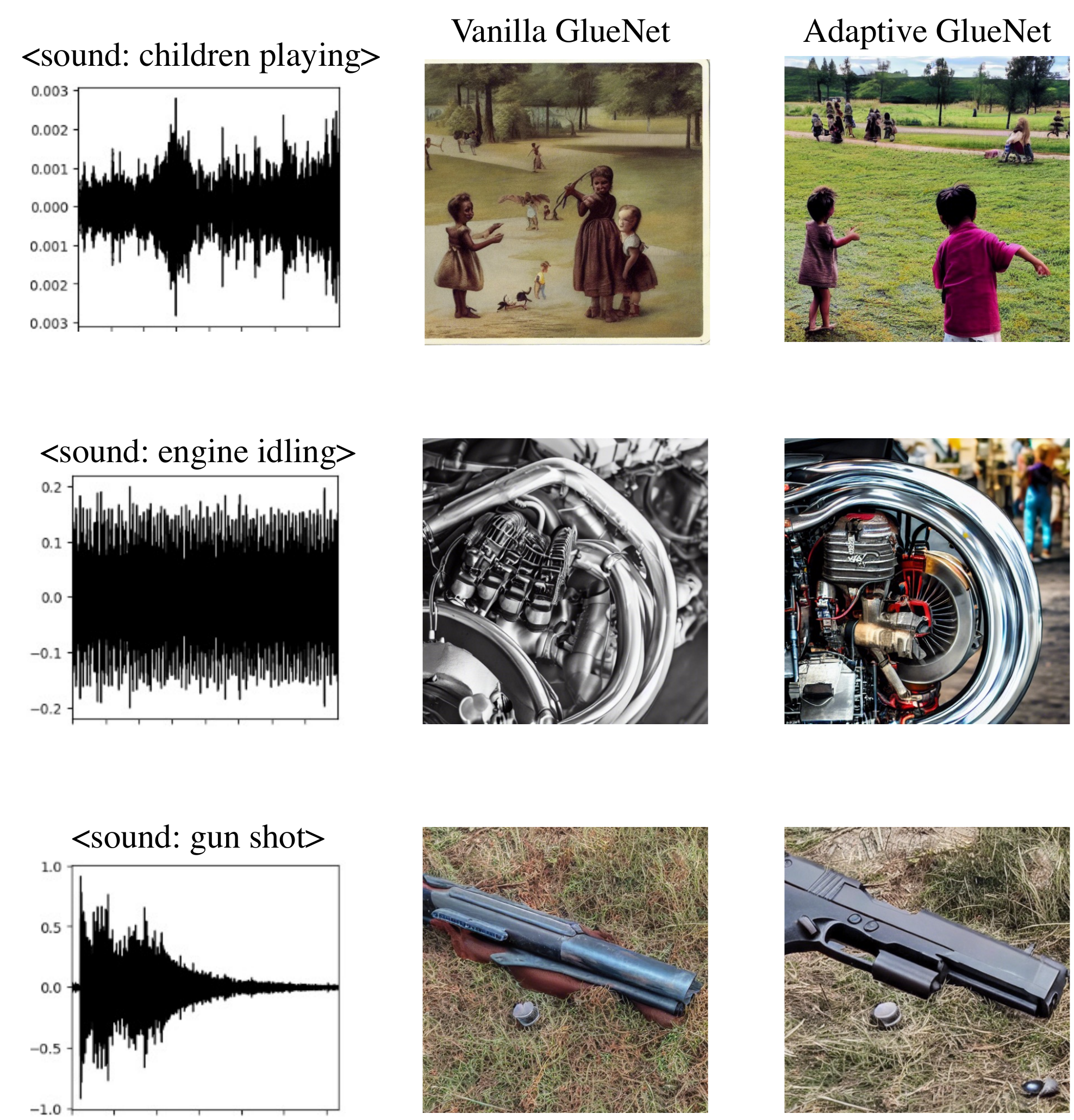}
\caption{Sound-to-image generation (Part 2/2) with vanilla and adaptive GlueNets.   }\label{fig:sound_to_image_2}
\end{figure}

We provide additional visual results in Figures~\ref{fig:sound_to_image_1} and ~\ref{fig:sound_to_image_2}, comparing the vanilla GlueNet (without the re-weight objective) with the Adaptive GluNet (with the re-weight objective) as described in Section \ref{sec:cross-modal} of the main paper. These figures clearly show that the Adaptive GluNet improves the stability and accuracy of the generated images compared to the vanilla GlueNet, thus demonstrating the effectiveness of the objective re-weighting technique proposed in Section \ref{sec:cross-modal}.

\section{Multimodal-to-Image Generation}
\begin{figure}[t]
\centering
\includegraphics[height=0.9\linewidth,width=0.75\linewidth]{  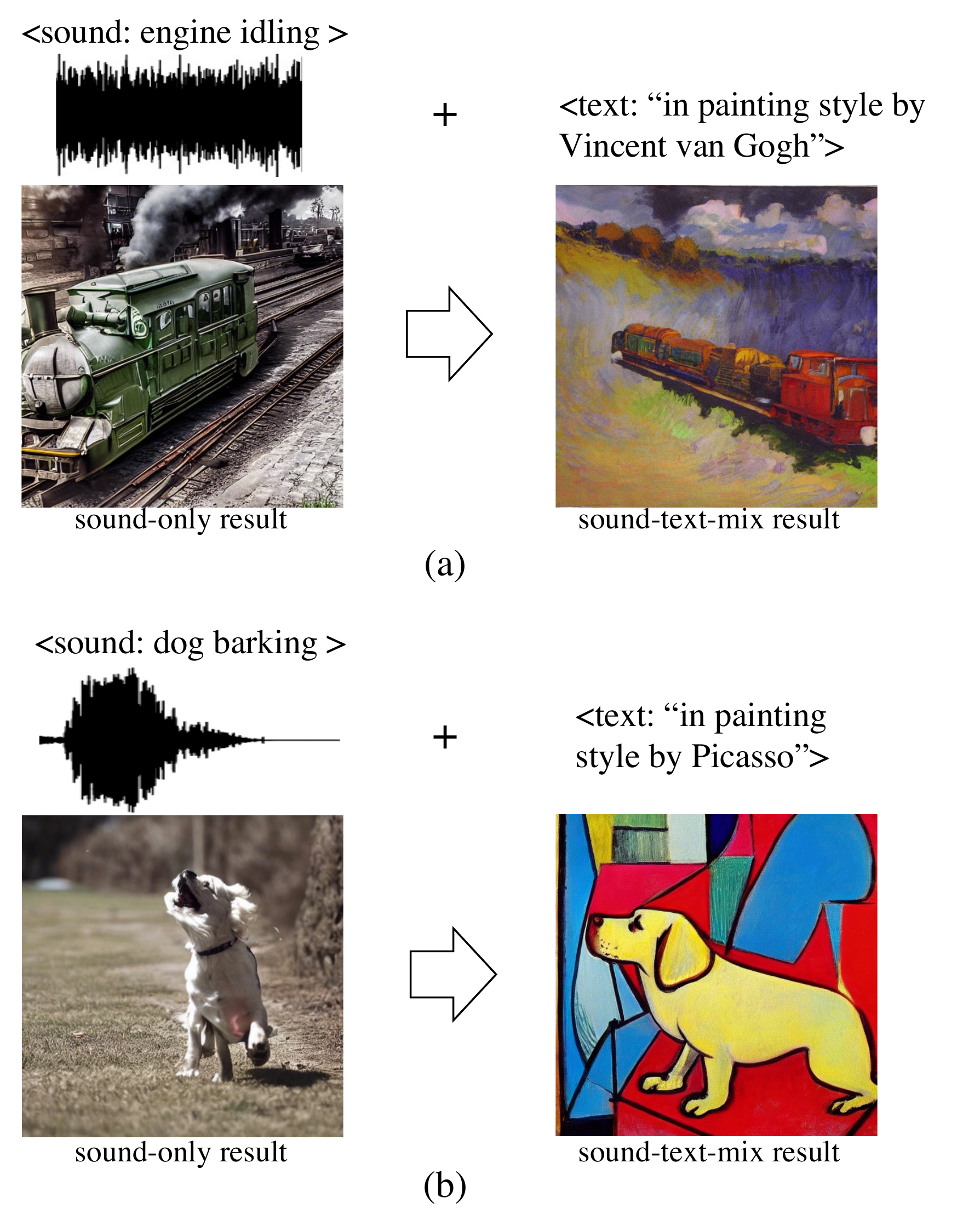}
\caption{Multimodal-to-image generation (sound-text-mix) results. The condition encoders are ClipText and AudioCLIP+GlueNet. The two-modality features are fused to Stable Diffusion Unet to generate such right-side result.  }\label{fig:sound_text_to_image}
\end{figure}

According to Section \ref{sec:cross-modal} of the main paper, the fusion of multi-modal condition features from different encoders can be achieved through non-parametric operations, such as concatenating the top K signals and averaging the rest (excluding the last K). More multi-modal generation results are presented in Figure~\ref{fig:sound_text_to_image}. In our experiment, we used a text encoder (CLIPText) to extract the text embedding (with the input of "in painting style by Picasso"). We also inputted audio data to the AudioCLIP model, which was appended with a GlueNet to map it into an embedding. Then, we applied the proposed feature fusion operator, merging both text and sound embeddings, which enabled the stable diffusion model to generate reasonable results.

\begin{figure}[t]
\centering
\includegraphics[height=0.625\linewidth,width=\linewidth]{  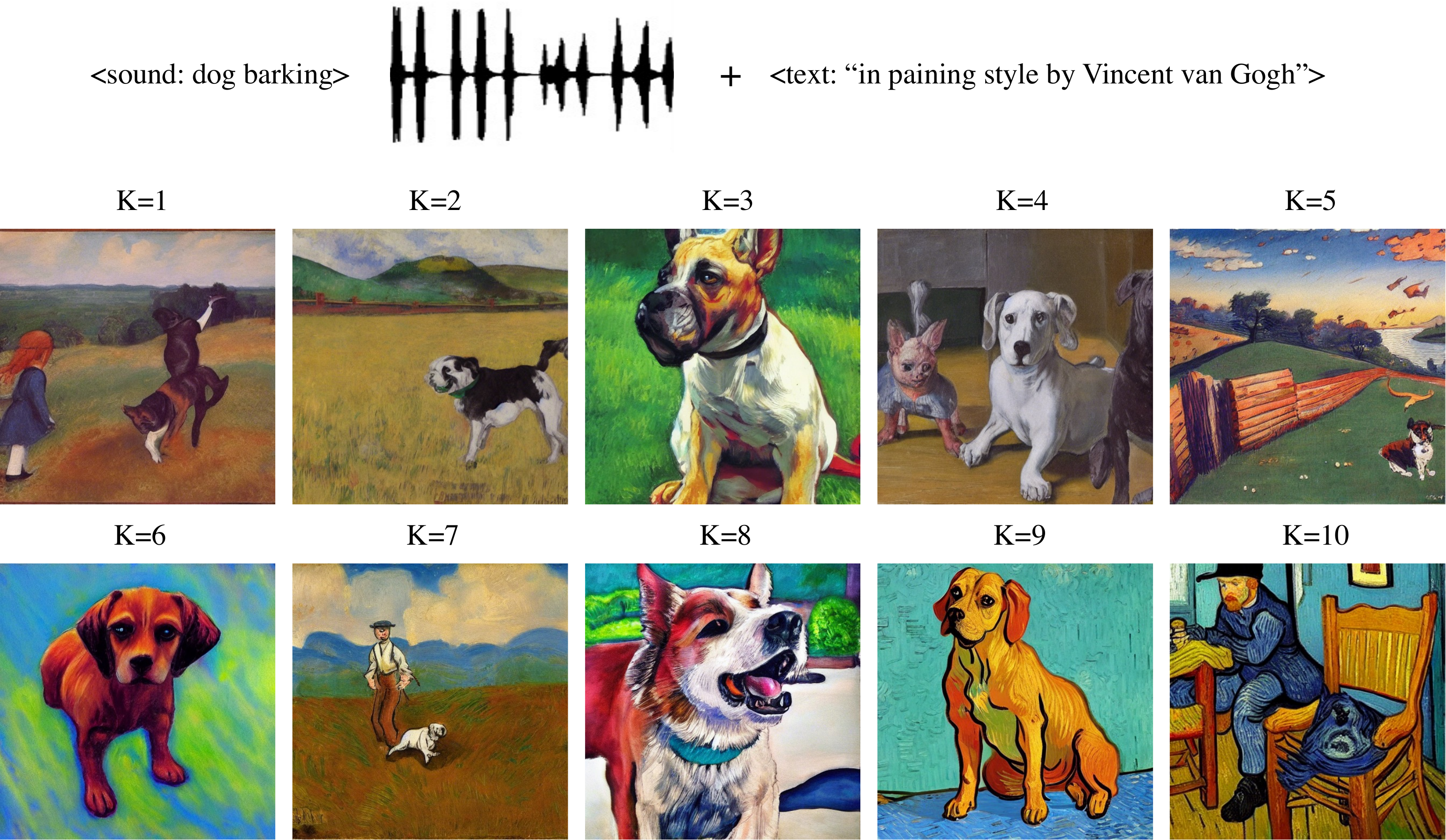}
\caption{Analysis of top $K$ selected signals for multimodal (sound and text) feature fusion. This operation is  non-parametric (also training-free) which only needs concatenating top  $K$ token signals and averaging the rest. }\label{fig:singal_k_analysis}
\end{figure}

We have conducted an ablation study to determine the optimal value of K for fusing multi-modal condition features using the non-parametric operations described in Section \ref{sec:cross-modal} of the main paper. The results, which are presented in Figure~\ref{fig:singal_k_analysis}, show that when $K<=6$, the audio signals dominate the generation process. As $K$ increases, text signals gradually begin to appear and eventually dominate the conditioning when $K=10$. Therefore, to strike a good balance between the two cross-modal signals, we recommend selecting an appropriate value of $K$ based on empirical observations.

\end{document}